%% file: main.tex
\documentclass[10pt,twocolumn,letterpaper]{article}

\usepackage[pagenumbers]{cvpr}
\input{math_commands.tex}


\usepackage[pagebackref,breaklinks,colorlinks,allcolors=cvprblue]{hyperref}
\usepackage{duckuments}
\usepackage{adjustbox}
\usepackage{array}
\usepackage{booktabs}
\usepackage{subcaption}
\usepackage{lipsum}
\usepackage{amsmath}
\usepackage{float}
\usepackage{multirow}

\usepackage{listings}
\usepackage{algorithm}
\usepackage{algpseudocode}
\usepackage{bibunits}
\usepackage{changepage}

\definecolor{DarkBlue}{rgb}{0,0.08,0.45}

\hypersetup{%
colorlinks=true,
linkcolor=DarkBlue,
citecolor=DarkBlue,
filecolor=DarkBlue,
urlcolor=DarkBlue}

\newcommand\blfootnote[1]{%
  \begingroup
  \renewcommand\thefootnote{}\footnote{#1}%
  \addtocounter{footnote}{-1}%
  \endgroup
}

\newcommand{\sname}{CoordTok\xspace}

\title{Efficient Long Video Tokenization via Coordinate-based Patch Reconstruction}

\author{Huiwon Jang$^{1}$
\qquad Sihyun Yu$^{1}$
\qquad Jinwoo Shin$^{1}$
\qquad Pieter Abbeel$^{2}$
\qquad Younggyo Seo$^{2}$\\
{$^1$KAIST \,\, $^2$UC Berkeley}
}

\begin{document}
\maketitle

\input{sec/0_abstract}
\input{sec/1_intro}
\input{sec/3_method}
\input{sec/4_experiments}
\input{sec/2_related_works}
\input{sec/5_conclusion}

\section*{Acknowledgements}
This work was supported by Institute for Information \& communications Technology Promotion(IITP) grant funded by the Korea government(MSIT) (No.RS-2019-II190075 Artificial Intelligence Graduate School Program(KAIST); No.RS-2021-II212068, Artificial Intelligence Innovation Hub) and Samsung Electronics Co., Ltd (IO201211-08107-01).
PA holds concurrent appointments as a Professor at UC Berkeley and as an Amazon Scholar. This paper describes work performed at UC Berkeley and is not associated with Amazon.
YS is supported in part by Multidisciplinary University Research Initiative (MURI) award by the Army Research Office (ARO) grant No. W911NF-23-1-0277.
We thank NVIDIA for providing
compute resources through the NVIDIA Academic DGX Grant.

{
    \small
    \bibliographystyle{ieeenat_fullname}
    \bibliography{main}
}

\input{sec/X_suppl}

\end{document}

%% file: math_commands.tex
\usepackage{amsmath,amsfonts,bm}

\def\eqref#1{equation~\ref{#1}}

\def\1{\bm{1}}

\def\rve{{\mathbf{e}}}

\def\rvh{{\mathbf{h}}}

\def\rvx{{\mathbf{x}}}

\def\rvz{{\mathbf{z}}}

\DeclareMathAlphabet{\mathsfit}{\encodingdefault}{\sfdefault}{m}{sl}
\SetMathAlphabet{\mathsfit}{bold}{\encodingdefault}{\sfdefault}{bx}{n}



%% file: sec/0_abstract.tex
\begin{abstract}
Efficient tokenization of videos remains a challenge in training vision models that can process long videos.
One promising direction is to develop a tokenizer that can encode long video clips, as it would enable the tokenizer to leverage the temporal coherence of videos better for tokenization.
However, training existing tokenizers on long videos often incurs a huge training cost as they are trained to reconstruct all the frames at once.
In this paper, we introduce CoordTok, a video tokenizer that learns a mapping from coordinate-based representations to the corresponding patches of input videos, inspired by recent advances in 3D generative models. 
In particular, CoordTok encodes a video into factorized triplane representations and reconstructs patches that correspond to randomly sampled $(x,y,t)$ coordinates.
This allows for training large tokenizer models directly on long videos without requiring excessive training resources.
Our experiments show that CoordTok can drastically reduce the number of tokens for encoding long video clips. 
For instance, CoordTok can encode a 128-frame video with 128$\times$128 resolution into 1280 tokens, while baselines need 6144 or 8192 tokens to achieve similar reconstruction quality.
We further show that this efficient video tokenization enables memory-efficient training of a diffusion transformer that can generate 128 frames at once.
\blfootnote{Project website: {
\href{https://huiwon-jang.github.io/coordtok/}{\nolinkurl{huiwon-jang.github.io/coordtok}}}}
\blfootnote{Correspondence to {\texttt{mail@younggyo.me}}.} 
\end{abstract}

%% file: sec/1_intro.tex
\section{Introduction}
\label{sec:intro}
\input{resource/motivation_arxiv}

Efficient tokenization of videos remains a challenge in developing vision models that can process long videos.
While recent video tokenizers have achieved higher compression ratios \citep{ge2022long,blattmann2023align,yu2023magvit,yu2023language,videoworldsimulators2024,wang2024emu3} compared to using image tokenizers for videos (\ie, frame-wise compression) \citep{zhou2024transfusion,team2024chameleon},
the vast scale of video data still requires us to design a more efficient video tokenizer.

One promising direction for efficient video tokenization is enabling video tokenizers to exploit the \textit{temporal coherence} of videos.
For instance, video codecs \citep{rijkse1996h,marpe2006h,sullivan2012overview,mukherjee2015technical} extensively utilize such coherence for video compression by extracting keyframes and encoding the difference between them.
In fact, there have been several recent works based on a similar intuition that train a tokenizer to encode videos into factorized representations \citep{yu2023video,kim2024hybrid,yu2024efficient}.
However, a key limitation is that existing tokenizers are typically trained to encode short video clips because of high training cost, but it is more likely that tokenizers can better exploit the temporal coherence when they are trained on longer videos. 
For instance, because tokenizers are trained to reconstruct all the frames at once, their training cost increases linearly with the length of videos (see \Cref{subfig:long-scalability}).
This makes it difficult to train tokenizers that can encode long videos and thus capture the temporal coherence of videos (see \Cref{subfig:inter-clip-inconsistency}).

In this paper, we aim to design a video tokenizer that can be easily scaled up to encode long videos.
To this end, we draw inspiration from recent works that have successfully trained large 3D generative models in a compute-efficient manner \citep{hong2023lrm,jun2023shap,miyato2023gta,liu2024one}.
Their key idea is to train a model that learns a mapping from randomly sampled $(x, y, z)$ coordinates to RGB and density values instead of training with all the possible coordinates at once.

In particular, we ask: can we utilize a similar idea to design a scalable video tokenizer?
Actually, there have been recent studies that formulate the video reconstruction as a problem of learning the mapping from $(x, y, t)$ coordinates to RGB values \citep{kim2022scalable,chen2022videoinr}.
However, they rather focus on compressing each individual video instead of training a video tokenizer that can encode a diverse set of videos.

We introduce CoordTok: \textbf{Coord}inate-based patch reconstruction for long video \textbf{Tok}enization, a scalable video tokenizer that learns a mapping from coordinate-based representations to the corresponding patches of input videos.
The key idea of CoordTok is to encode a video into factorized triplane representations \citep{kim2022scalable,yu2023video} and reconstruct patches that correspond to randomly sampled $(x,y,t)$ coordinates (see \Cref{fig:method_overview}).
This enables the training of large tokenizers directly on long videos without excessive memory and computational requirements (see \Cref{subfig:long-scalability}).

To investigate whether training a video tokenizer on long video clips indeed leads to more efficient tokenization,
we compare CoordTok with other baselines \citep{ge2022long,yu2023video,yu2023magvit,wang2024larp} on the UCF-101 dataset \citep{soomro2012ucf101}.
Our experiments show that, by exploiting the temporal coherence of videos,
CoordTok significantly reduces the number of tokens for encoding long videos compared to baselines.
For instance, CoordTok encodes a 128-frame video with 128$\times$128 resolution into only 1280 tokens, while baselines require 6144 or 8192 tokens to achieve similar encoding quality.
We also show that efficient tokenization with CoordTok enables 
memory-efficient training of a diffusion transformer \citep{peebles2023scalable,ma2024sit} that can generate a 128-frame video at once.
Finally, we provide an extensive analysis on the effect of various design choices.
\input{resource/method_overview}

\vspace{0.02in}
\noindent We summarize the contributions of this paper below:
\vspace{0.02in}
\begin{itemize}[topsep=0.0pt,itemsep=1.2pt]
    \item We introduce CoordTok, a scalable video tokenizer that learns a mapping from coordinate-based representations to the corresponding patches of input videos.
    \item We show that CoordTok can leverage the temporal coherence of videos for tokenization, drastically reducing the number of tokens required for encoding long videos.
    \item We show that efficient video tokenization with CoordTok enables memory-efficient training of a diffusion transformer \citep{peebles2023scalable,ma2024sit} that can generate long videos at once.
\end{itemize}

%% file: resource/motivation_arxiv.tex
\begin{figure*}[t]
\centering
\begin{subfigure}{.37\textwidth}
\centering
\includegraphics[width=0.95\textwidth]{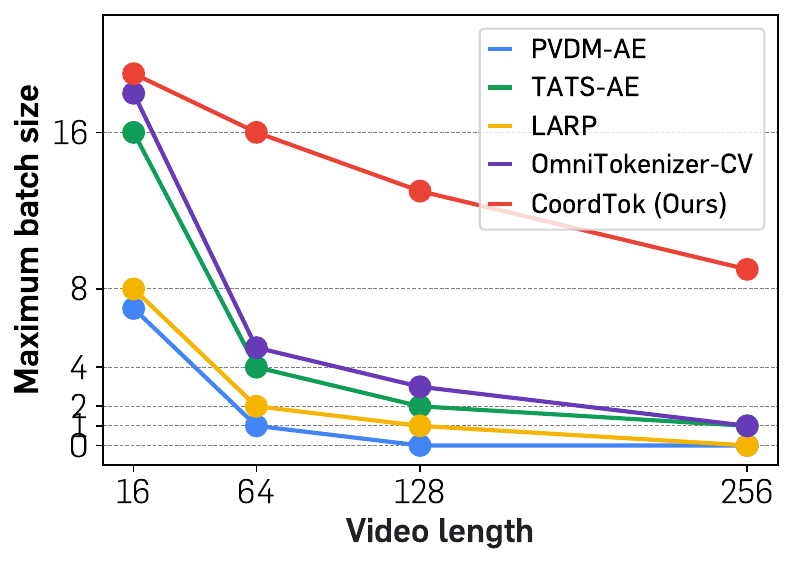}
\caption{\textbf{Maximum batch-size when training video tokenizers} on 128$\times$128 resolution videos with varying lengths, measured with a single NVIDIA 4090 24GB GPU.}
\label{subfig:long-scalability}
\end{subfigure}
\hfill
\begin{subfigure}{.61\textwidth}
\centering
\includegraphics[width=\textwidth]{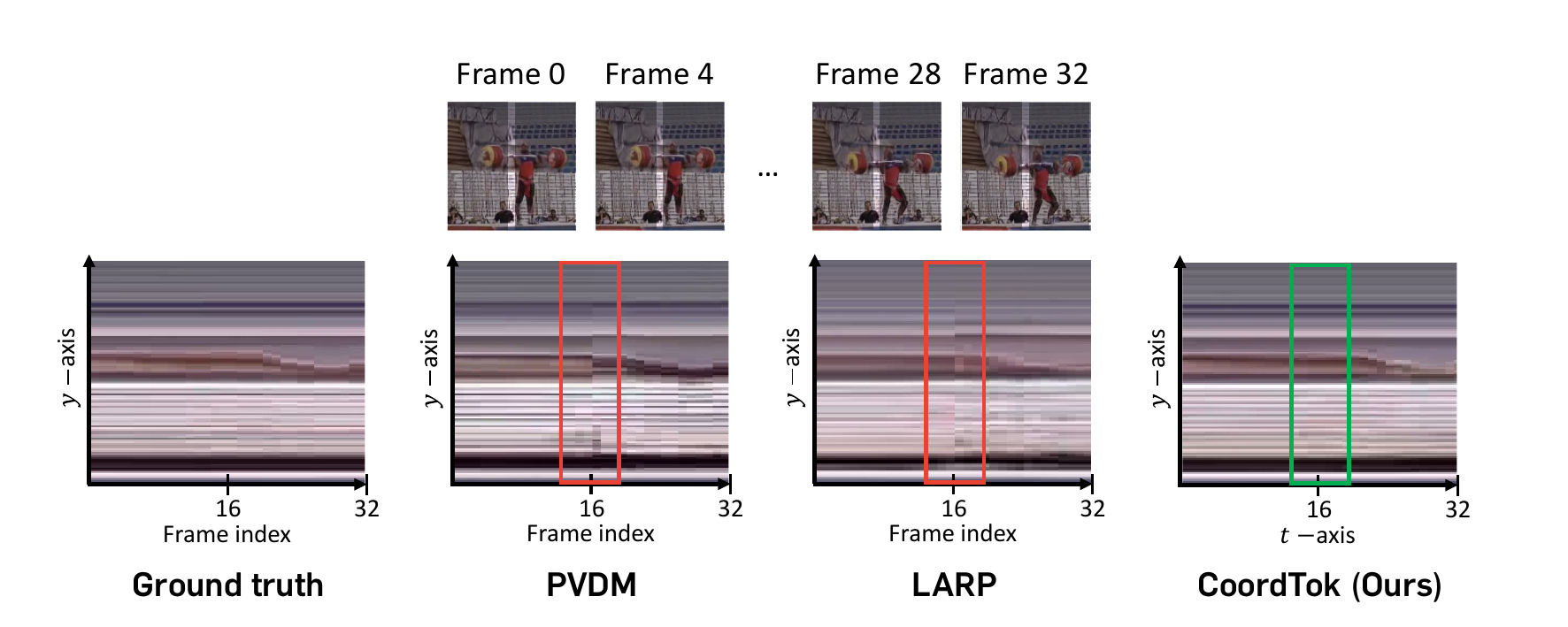}
\caption{\textbf{Inter-clip reconstruction consistency of video tokenizers.} Existing video tokenizers \citep{ge2022long,yu2023video,wang2024larp} show the pixel-value inconsistency between short clips (16 frames). In contrast, Our tokenizer shows the temporally consistent reconstruction.}
\label{subfig:inter-clip-inconsistency}
\end{subfigure}
\vspace{-0.07in}
\caption{\textbf{Limitation of existing video tokenizers.} 
(a) Existing video tokenizers \citep{ge2022long,yu2023video,wang2024larp} are often not scalable to long videos because of excessive memory and computational demands.
This is because they are trained to reconstruct all video frames at once, \ie, a giant 3D array of pixels, which incurs a huge computation and memory burden in training especially when trained on long videos.
For instance, PVDM-AE \citep{yu2023video} becomes out-of-memory when trained to encode 128-frame videos when using a single NVIDIA 4090 24GB GPU.
(b) As a result, existing tokenizers are typically trained to encode up to 16-frame videos and struggle to capture the temporal coherence of videos.}
\label{fig:motivation}
\end{figure*}

%% file: resource/method_overview.tex
\begin{figure*}[tb]
\centering
\includegraphics[width=0.99\linewidth]{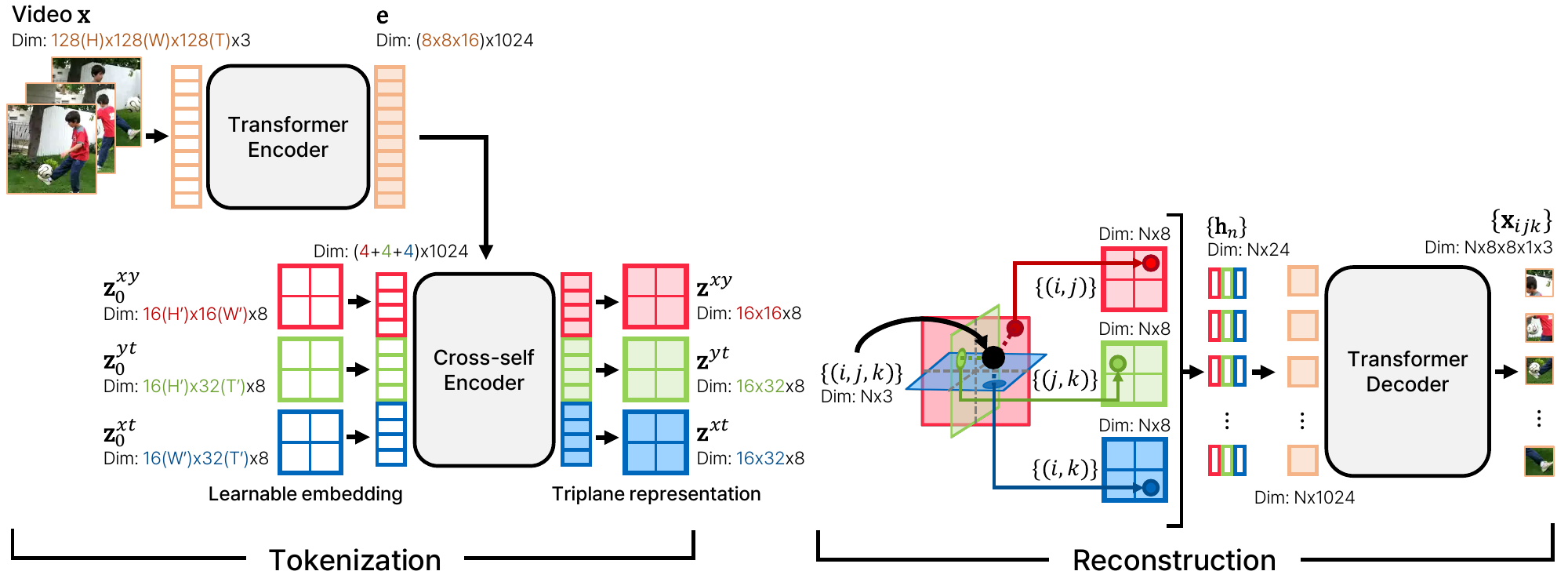}
\caption{\textbf{Overview of CoordTok.} We design our encoder to encode a video $\rvx$ into factorized triplane representations $\rvz = [\rvz^{xy}, \rvz^{yt}, \rvz^{xt}]$ which can efficiently represent the video with three 2D latent planes.
Given the triplane representations $\mathbf{z}$, our decoder learns a mapping from $(x,y,t)$ coordinates to RGB pixels within the corresponding patches.
In particular, we extract coordinate-based representations of $N$ sampled coordinates by querying the coordinates from triplane representations via bilinear interpolation. 
Then the decoder aggregates and fuses information from different coordinates with self-attention layers and project outputs into corresponding patches.
This design enables us to train tokenizers on long videos in a compute-efficient manner by avoiding reconstruction of entire frames at once.
}
\label{fig:method_overview}
\end{figure*}

%% file: sec/3_method.tex
\section{Method}

In this section, we present CoordTok, a scalable video tokenizer that can efficiently encode long videos.
In a nutshell, CoordTok encodes a video into factorized triplane representations \citep{kim2022scalable,yu2023video} and learns a mapping from randomly sampled $(x, y, t)$ coordinates to pixels from the corresponding patches.
We provide the overview of CoordTok in \Cref{fig:method_overview}.

\paragraph{Problem setup} Let $\rvx$ be a video and $\mathcal{D}$ be a dataset consisting of videos.
Our goal is to train a video tokenizer that encodes a video $\rvx \in \mathcal{D}$ into tokens (or a low-dimensional latent vector) $\rvz$ and decodes $\rvz$ into $\rvx$.
In particular, we want the tokenizer to be \textit{efficient} so that it can encode videos into fewer number of tokens as possible but still can decode tokens to the original video $\rvx$ without loss of information.

\subsection{Encoder}
\label{sec:encoder}
Given a video $\rvx$, we divide the video into non-overlapping space-time patches. 
We then add learnable positional embeddings and process them through a series of transformer layers \citep{vaswani2017attention} to obtain video features $\rve$.

After that, we encode video features $\rve$ into factorized triplane representations \citep{chan2022efficient,yu2023video}, \ie, $\rvz = [\rvz^{xy}, \rvz^{yt}, \rvz^{xt}]$, where the planes have the shape of $H' \times W'$, $W' \times T'$, and $H' \times T'$, respectively. Intuitively, $\rvz^{xy}$ captures the global content in $\rvx$ across time (\eg, layout and appearance of the scene or object), $\rvz^{yt}$ and $\rvz^{xt}$ capture the underlying motion in $\rvx$ across two spatial axes (see \Cref{fig:latent-visualization} for visualization).
This design is efficient because it represents a video with three 2D latent planes instead of 3D latents widely used in prior approaches \citep{ge2022long,yu2023magvit,wang2024emu3}.

We implement our encoder based on the memory-efficient design of a recent 3D generation work \citep{hong2023lrm} that introduces learnable embeddings and translates them to triplane representations.
Specifically, we first introduce learnable embeddings $\rvz_{0} = [\rvz_{0}^{xy}, \rvz_{0}^{yt}, \rvz_{0}^{xt}]$.
We then process them through a series of cross-self attention layers, where each layer consists of (i) cross-attention layer that attends to the video features $\rve$ and (ii) self-attention layer that attends to its own features.
In practice, we split each learnable embedding into four smaller equal-sized embeddings.
We then use them as inputs to the cross-self encoder, because we find it helps the model to use more computation by increasing the length of input sequence.
Finally, we project the outputs into triplane representations to obtain $\rvz = [\rvz^{xy}, \rvz^{yt}, \rvz^{xt}]$.

\subsection{Decoder}
Given the triplane representation $\rvz = [\rvz^{xy}, \rvz^{yt}, \rvz^{xt}]$, we implement our decoder to reconstruct partial video during the training stage by learning a mapping from $(i,j,k)$ coordinate to the pixels of the corresponding patch.

\paragraph{Input and target}
We use patch coordinates as inputs to the decoder and their corresponding patch RGB values as targets. Specifically, we first divide the video $\rvx$ into non-overlapping space-time patches. We note that the configuration of patches, \eg, patch sizes, may differ from the one used in the video encoder. We then convert each patch index into the $(i,j,k)$ coordinates representing the center position of the patch along each $x$, $y$, and $t$ axis relative to the entire video $\rvx$, where $i,j,k \in [0, 1]$. Finally, we randomly sample $N$ patches.
We find that sampling only $3$\% of video patches can achieve strong performance (see \Cref{tab:analysis-partial-recon} for the effect of sampling).
\begin{gather}
\begin{aligned}
&\text{Input:} && [(i_1, j_1, k_1), \cdots, (i_N, j_N, k_N)] \\
&\text{Target:} && [\rvx_{i_1 j_1 k_1}, \cdots, \rvx_{i_N j_N k_N}]
\label{eq:decoder_input_output}
\end{aligned}
\end{gather}

\paragraph{Coordinate-based representations}
\input{resource/qualitative_results}
\input{resource/main_results_fig}

As inputs to the transformer decoder, we use coordinate-based representations $\rvh$ that are obtained by querying each input coordinate from triplane representation via bilinear interpolation. Specifically, let $(i, j, k)$ be one of sampled coordinates.
We extract $\rvh^{xy}$ by querying $(i, j)$ from $\rvz^{xy}$, $\rvh^{yt}$ by querying $(j, k)$ from $\rvz^{yt}$, and $\rvh^{xt}$ by querying $(i, k)$ from $\rvz^{xt}$. More specifically, let $(l, m, n)$ be the indices in the triplane representation corresponding to $(i, j, k)$, obtained using the floor function, \ie, $(l,m,n) = (\lfloor i H' \rfloor, \lfloor jW' \rfloor, \lfloor k T' \rfloor)$. Then, coordinate-based representations are computed as follows:
\vspace{-0.07in}
\begin{adjustwidth}{-0.15cm}{0pt}
\begin{gather}
\begin{aligned}
&\rvh^{xy} = \text{Bilerp}((i,j); \rvz^{xy}_{lm}, \rvz^{xy}_{l,m+1},\rvz^{xy}_{l+1,m},\rvz^{xy}_{l+1,m+1}) \\
&\rvh^{yt} = \text{Bilerp}((j,k); \rvz^{yt}_{mn}, \rvz^{yt}_{m,n+1},\rvz^{yt}_{m+1,n},\rvz^{yt}_{m+1,n+1}) \\
&\rvh^{xt} = \text{Bilerp}((i,k); \rvz^{xt}_{ln}, \rvz^{xt}_{l,n+1},\rvz^{xt}_{l+1,n},\rvz^{xt}_{l+1,n+1})
\label{eq:bilinear_interpolation}
\end{aligned}
\end{gather}
\end{adjustwidth}
where $(\rvz^{xy}_{lm}, \rvz^{yt}_{mn}, \rvz^{xt}_{ln})$ indicates the latent vector in $\rvz$ at indices $(l,m,n)$, and $\text{Bilerp}(\cdot;\cdot)$ is the bilinear interpolation operation at the input coordinate between given vectors.
We then concatenate them to get the coordinate-based representation of $(i,j,k)$, \ie, $\rvh:=\text{Concat}(\rvh^{xy}, \rvh^{yt}, \rvh^{xt})$.

\paragraph{Patch reconstruction}
Given $N$ coordinate-based representations $[\rvh_{1}, ..., \rvh_{N}]$, our decoder processes them through a series of self-attention layers, enabling each $\rvh_{n}$ to attend to other representations $\rvh_{m}$.
This allows the decoder to aggregate and fuse the information from different coordinates.
We then use a linear projection layer to process the output from each $\rvh_{n}$ to pixels of the corresponding patch $\rvx_{i_n j_n k_n}$.
Finally, we update the parameters of our encoder and decoder to minimize an $\ell_{2}$ loss between the reconstructed pixels and original pixels.

To further improve the quality of reconstructed videos, we introduce an additional fine-tuning phase where we train our tokenizer with both $\ell_{2}$ loss and LPIPS loss \citep{zhang2018unreasonable}.
Specifically, instead of sampling coordinates, we randomly sample a few frames and use all coordinates within the sampled frames for fine-tuning.
This enables the tokenizer to compute and minimize LPIPS loss, which requires reconstructing the entire frame.
While we find that sampling frames instead of coordinates from the beginning of the training is harmful due to the lack of diversity in training data (see \Cref{tab:analysis-partial-recon}), we find that fine-tuning with sampled frames improves the quality of reconstructed videos.

%% file: resource/qualitative_results.tex
\begin{figure*}[t]
\centering
\includegraphics[width=0.95\linewidth]{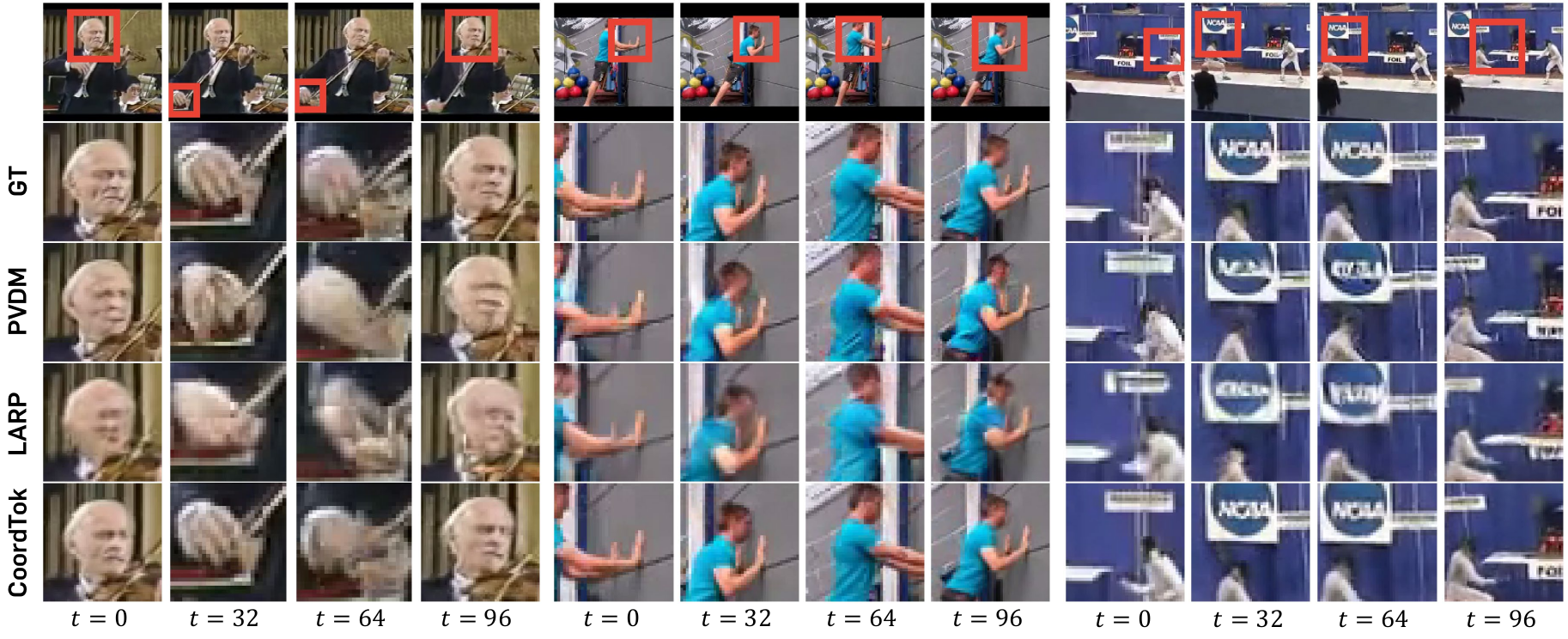}
\vspace{-0.1in}
\caption{\textbf{128-frame, 128$\times$128 resolution video reconstruction results} from \sname (Ours) and baselines \citep{yu2023video,wang2024larp} trained on the UCF-101 dataset \citep{soomro2012ucf101}.
For each frame, we visualize the ground-truth (GT) and reconstructed pixels within the region highlighted in the red box, where CoordTok achieves noticeably better reconstruction quality than other baselines.
}
\vspace{-0.1in}
\label{fig:qualitative_results}
\end{figure*}

%% file: resource/main_results_fig.tex
\begin{figure}[t]
\vspace{-0.01in}
\centering\small
    \includegraphics[width=0.9\linewidth]{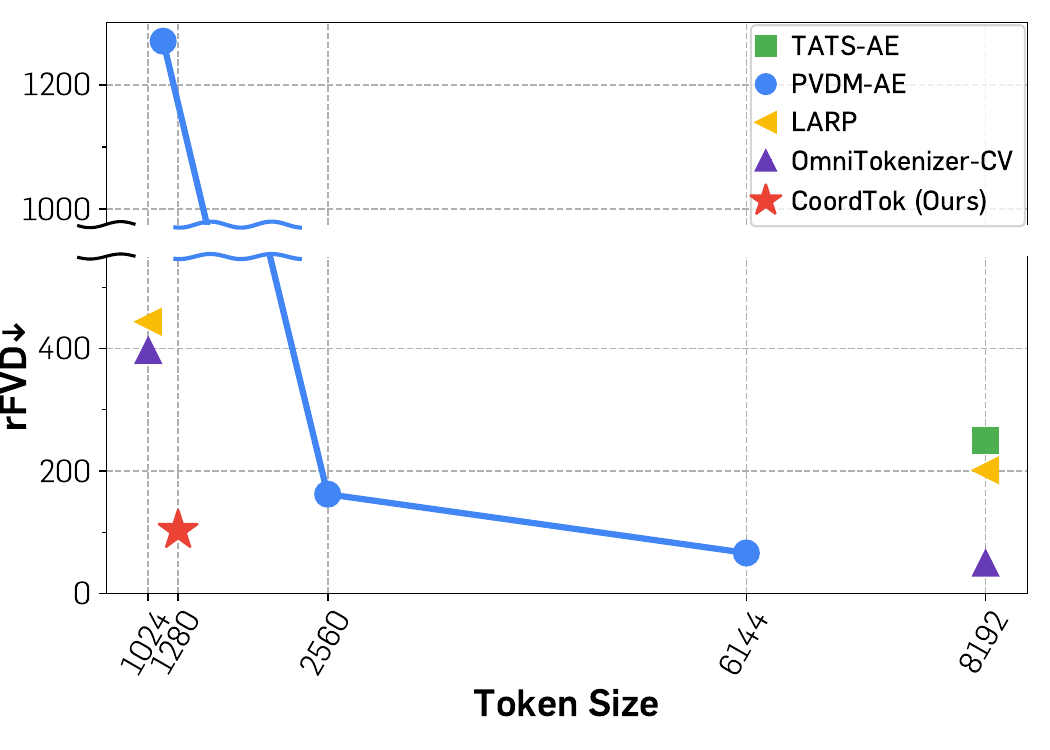}
    \vspace{-0.07in}
    \caption{\textbf{CoordTok can efficiently encode long videos.} rFVD scores of video tokenizers, evaluated on 128-frame videos, with respect to the token size. $\downarrow$ indicates lower values are better.
    }
    \label{fig:recon_vs_token}
\end{figure}

%% file: sec/4_experiments.tex
\section{Experiments}
We design experiments to investigate following questions:
\vspace{0.02in}
\begin{itemize}[leftmargin=5mm,itemsep=0mm]
    \item Can \sname efficiently encode long videos? Does encoding long videos lead to efficient video tokenization (\Cref{fig:qualitative_results,fig:recon_vs_token,tab:main})
    \item Can \sname learn meaningful tokens that can be used for downstream tasks such as video generation? (\Cref{tab:long-video-generation}) Can efficient video tokenization improve video generation models? (\Cref{tab:long-video-generation-efficiency})
    \item What is the effect of various design choices? (\Cref{fig:analysis-scalability,tab:analysis-partial-recon})
\end{itemize}

\subsection{Experimental Setup}
\paragraph{Implementation details}
We conduct all our experiments on the UCF-101 \citep{soomro2012ucf101} dataset.
Following the setup of prior works \citep{ge2022long,yu2023magvit}, we use the train split of the UCF-101 dataset for training.
For preprocessing videos, we resize and center-crop the frames to $128\times128$ resolution.
We train our tokenizer using the AdamW optimizer \citep{loshchilov2017decoupled} with a batch size of 256, where each sample is a randomly sampled 128-frame video.
We use $N=1024$ coordinates for main training and $N=4096$ for fine-tuning.
For the main experimental results, we train CoordTok for 1M iterations and further fine-tune it with LPIPS loss for 50k iterations.
For analysis and ablation studies, we train CoordTok for 200k iterations and further fine-tune it with LPIPS loss for 10k iterations.
For model configurations such as embedding dimension and number of layers, we mostly follow the architectures of vision transformers (ViTs; \citep{dosovitskiy2020image}).
We provide more detailed implementation details in \cref{appendix:implementation_details}.

\input{resource/main_results}

\paragraph{Evaluation}
For evaluating the quality of reconstructed videos, we follow the setup of MAGVIT \citep{yu2023magvit} that reports reconstruction Fr\'echet video distance (rFVD; \citep{unterthiner2019fvd}), peak signal-to-noise ratio (PSNR), LPIPS \citep{zhang2018unreasonable}, and SSIM \citep{wang2004image}.
We use 10000 video clips of length 128 for evaluation.
For evaluating the quality of generated videos, we follow the setup of StyleGAN-V \citep{skorokhodov2022stylegan} that reports FVD measured with 2048 video clips.
We provide more details of evaluation metrics in \cref{appendix:evaluation_details}.

\input{resource/generation_results_single_row}
\input{resource/analysis_scalability}

\subsection{Long video tokenization}

\paragraph{Setup}
To investigate whether training CoordTok to encode long videos at once leads to efficient tokenization, we consider a setup where tokenizers encode 128-frame videos.
Because existing tokenizers cannot encode such long videos at once,
we split videos into multiple 16-frame video clips, use baseline tokenizers to encode each of them, and then concatenate the tokens from entire splits.
For CoordTok, we train our tokenizer to encode 128-frame videos at once.
We provide more details in \cref{appendix:implementation_details}.

\paragraph{Baselines}
We mainly consider tokenizers used in recent image or video generation models as our baselines.
We first consider MaskGIT-AE \citep{chang2022maskgit}, an image tokenizer, as our baseline to evaluate the benefit of using video tokenizers for encoding videos.
Moreover, we consider PVDM-AE \citep{yu2023video}, which encodes a video into factorized triplane representations and decodes all frames at once, as another baseline.
Comparison with PVDM-AE enables us to evaluate the benefit of our decoder design because it shares the same latent structure with CoordTok.
We further consider recent video tokenizers that encode videos into 3D latents, \ie, TATS-AE \citep{ge2022long}, MAGVIT-AE-L \citep{yu2023magvit}, LARP \citep{wang2024larp}, OmniTokenizer-DV \citep{wang2024omnitokenizer}, and OmniTokenizer-CV \citep{wang2024omnitokenizer} as our baselines. For a fair comparison, we train all baselines from scratch on UCF-101 or use the model weights trained on UCF-101 following their official implementations.
In addition, we compare \sname to CosmosTokenizer-CV \citep{agarwal2025cosmos}, a state-of-the-art tokenizer, although it is not a directly comparable baseline because it is trained on a large-scale dataset.
We provide more details of each baseline in \cref{appendix:baselines}.

\paragraph{Results}
For qualitative evaluation, we provide videos reconstructed by CoordTok and other baseline tokenizers in \cref{fig:qualitative_results}.
Notably, we find that CoordTok efficiently encodes 128-frame videos into only 1280 tokens.
In contrast, baselines achieve significantly worse reconstruction quality when they use a similar number of tokens to CoordTok.
For instance, CoordTok can encode 128-frame videos to 1280 tokens with a rFVD score of 103, while PVDM-AE achieves $>$1000 rFVD score when using 1152 tokens.
This highlights the benefit of our decoder design, which enables the tokenizer to exploit the temporal coherence of long videos better for efficient tokenization.
Moreover, \cref{tab:main} shows CoordTok outperforms baseline tokenizers across diverse metrics that assess the quality of reconstructed frames.

\subsection{Long video generation} \label{sec:long_video_generation}
\paragraph{Setup}
To investigate whether CoordTok can encode long videos into meaningful tokens,
we consider an unconditional video generation setup where we train a model to produce 128-frame videos.
Videos of length 128 are often considered too long to be generated at once, so several works use techniques such as iterative generation \citep{yu2023video} for generating long videos.
However, because CoordTok can efficiently encode long videos, we train our model to generate 128-frame videos at once.
Specifically, we encode 128-frame videos into 1280 tokens with CoordTok and train a SiT-L/2 model \citep{ma2024sit}, a recent flow-based transformer model, for 600K iterations with a batch size of 64.
We then use the model to generate 128-frame videos using the Euler-Maruyama sampler with 250 sampling steps.
We provide more implementation details in \cref{appendix:implementation_details}.

\paragraph{Baselines}
We consider recent video generation models that can generate 128-frame videos as baselines, \ie, MoCoGAN \citep{tulyakov2018mocogan}, MoCoGAN-HD \citep{tian2021good}, DIGAN \citep{yu2022generating}, StyleGAN-V \citep{skorokhodov2022stylegan}, PVDM-L \citep{yu2023video}, HVDM \citep{kim2024hybrid}, and Latte-L/2 \citep{ma2024latte_rebuttal}.
We provide more details of each baseline in \cref{appendix:baselines}.

\paragraph{Results}
\Cref{tab:long-video-generation} provides the quantitative evaluation of our model, \ie, CoordTok-SiT-L/2, and other video generation models.
We find that CoordTok-SiT-L/2 achieves the best FVD score of 369.3, outperforming previous baselines.
This is an intriguing result considering that CoordTok-SiT-L/2 can generate 128-frame videos much faster than other baselines, as shown in \cref{tab:long-video-generation-efficiency}.
Moreover, to investigate whether efficient video tokenization improves video generation,
we evaluate the FVD scores of SiT-L/2 models trained with CoordTok using token sizes of 1280 and 3072.
\cref{fig:generation_fvd_curve} shows that SiT-L/2 trained with the token size of 1280 achieves consistently low FVD scores, even though there is no significant difference in the reconstruction quality of CoordTok with 1280 and 3072 tokens (see \cref{appendix:additional_analysis}).
This is likely because the reduced number of tokens makes it easier to train the SiT model.
For qualitative evaluation, we provide videos from CoordTok-SiT-L/2 in \Cref{appendix:additional_qualitative_results}.

\input{resource/analysis_recon_vs}

\subsection{Analysis and ablation studies}
\paragraph{Effect of model Size}
In \cref{subfig:analysis-model-size}, we investigate the scalability of CoordTok with respect to model sizes.
We evaluate three variants of CoordTok: CoordTok-S, CoordTok-B, and CoordTok-L.
Each variant has a different size for the encoder and decoder (see \cref{appendix:implementation_details} for detailed model configurations).
We find that the quality of reconstructed videos improves as the model size increases.
For instance, CoordTok-B achieves a PSNR of 25.2 while CoordTok-L achieves a PSNR of 26.9.

\paragraph{Effect of triplane size}
In \cref{subfig:analysis-spatial} and \cref{subfig:analysis-temporal}, we investigate the effect of spatial and temporal dimensions in triplane representations.
We evaluate CoordTok with varying spatial dimensions (16$\times$16, 32$\times$32, and 64$\times$64), and varying temporal dimensions (8, 16, and 32).
In general, we find that using larger planes improves the quality of reconstructed videos, as the model can better represent details within videos using more tokens.
This result suggests there is a trade-off between the number of tokens and the reconstruction quality.
In practice, we find reducing the spatial dimensions to 16$\times$16 while using a high temporal dimension of 32 strikes a good balance, achieving good quality of reconstructed videos with a relatively low number of tokens.

\paragraph{Effect of triplane representations}
We now examine the effect of one of our key design choices: encoding videos into triplane representations rather than 3D latents.
We hypothesize that CoordTok may struggle to encode dynamic videos, as decomposing a video to its content ($\rvz^{xy}$) and motion components ($\rvz^{yt}$, $\rvz^{xt}$) becomes difficult.
To investigate this, in \cref{subfig:recon_frame_distance}, we provide a scatter plot where the x-axis represents a metric for video dynamics and the $y$ axis represents the PSNR score.
As a metric for video dynamics, we use the mean $\ell_2$-distance between pixel values of consecutive frames (see \cref{appendix:evaluation_details} for more details).
As expected, we find that the correlation between reconstruction quality and the magnitude of dynamics is strong (-0.87) for CoordTok, compared to the weaker correlations for TATS-AE (-0.40) and MaskGIT-AE (-0.59), both of which use 3D latent structures.
This is one of the limitations of CoordTok, and addressing this by adopting techniques from video codecs, such as introducing multiple keyframes, could be an interesting future direction.

\paragraph{Effect of coordinate-based representations}
We further examine the effect of our design that trains using coordinate-based representations.
Our hypothesis is that the reconstruction quality of CoordTok is less sensitive to how fine-grained each video is, because CoordTok learns a mapping from each coordinate to pixels.
To investigate this, we measure the correlation between the PSNR score and a frequency metric proposed in \citet{yan2024elastictok} that utilizes a Sobel edge detection filter, where a larger frequency magnitude indicates a finer-grained video (see \cref{appendix:evaluation_details} for details).
As shown in \cref{subfig:recon_frequency}, the correlation between reconstruction quality and the frequency metric is weak (-0.37) for CoordTok, compared to stronger correlations for TATS-AE (-0.85) and MaskGIT-AE (-0.75).

\input{resource/analysis_partial_recon}
\input{resource/latent_visualization}
\paragraph{Effect of sampling}
We investigate two coordinate sampling schemes: (i) Random patch, which uses center coordinates of randomly sampled patches, and (ii) Random frame, which uses all coordinates from a few randomly sampled frames.
As shown in \cref{tab:analysis-partial-recon}, Random patch outperforms Random frame when sampling the same number of coordinates.
We hypothesize this is because Random frame fails to provide the tokenizer with sufficiently diverse training data.
For instance, sampling 3.125\% of video patches corresponds to sampling only 4 frames out of 128 in the Random frame scheme.
In contrast, Random patch uniformly samples patches from all 128 frames, which helps provide more diverse training data.
For Random patch, we find that sampling fewer coordinates reduces the training memory requirement but also degrades performance.

%% file: resource/main_results.tex
\begin{table*}[t]
\caption{\textbf{Reconstruction quality of image and video tokenizers.} We report metrics that measure the quality of reconstructed videos: PSNR, LPIPS, SSIM, and rFVD, computed using the 128$\times$128 resolution videos reconstructed by image and video tokenizers evaluated on the UCF-101 dataset \citep{soomro2012ucf101}. All models 
except CosmosTokenizer$^*$ \citep{agarwal2025cosmos} 
are trained on UCF-101.
Total \# tokens denotes the number of tokens required for encoding 128-frame videos.
\# Frames denotes number of frames in a video used for training tokenizers. $\downarrow$ and $\uparrow$ denotes whether lower or higher values are better, respectively.
}
\centering\small
\vspace{-0.1in}
\begin{adjustbox}{max width=0.8\textwidth}
    \begin{tabular}{l ccc cccc}
        \toprule
        & & & & \multicolumn{4}{c}{Reconstruction quality} \\
        \cmidrule(lr){5-8}
        Method & Token type & Total \# tokens &  \# Frames & PSNR$\uparrow$ & LPIPS$\downarrow$ & SSIM$\uparrow$ & rFVD$\downarrow$\\
        \midrule
        MaskGIT-AE \citep{chang2022maskgit} & Discrete & 8192 & 1 & 21.4 & 0.139 & 0.667 & \phantom{0}447.1 \\
        TATS-AE \citep{ge2022long}    & Discrete & 8192 & 16 & 23.2 & 0.213 & 0.792 & \phantom{0}249.4 \\
        MAGVIT-AE-L \citep{yu2023magvit} & Discrete & 8192 & 16 & 21.8 & 0.113 & 0.690 & -\\
        LARP \citep{wang2024larp} & Discrete & 8192 & 16 & 24.3 & 0.142 & 0.806 & \phantom{0}201.3 \\
        OmniTokenizer-DV \citep{wang2024omnitokenizer} & Discrete & 8192 & 17 & 26.1 & 0.113 & 0.871 & \phantom{0}\phantom{0}97.9 \\
        PVDM-AE \citep{yu2023video} & Continuous & 6144 & 16 & 26.5 & 0.120 & 0.859 & \phantom{0}\phantom{0}66.5 \\
        OmniTokenizer-CV \citep{wang2024omnitokenizer} & Continuous & 8192 & 17 & 28.3 & 0.081 & 0.913 & \phantom{0}\phantom{0}49.5 \\
        CosmosTokenizer-CV$^*$ \citep{agarwal2025cosmos} & Continuous & 8192 & 17 & 28.5 & 0.119 & 0.905 & \phantom{0}\phantom{0}87.8 \\
        \midrule
        LARP \citep{wang2024larp} & Discrete & 1024 & 16 & 22.0 & 0.181 & 0.766 & \phantom{0}443.5 \\
        OmniTokenizer-DV \citep{wang2024omnitokenizer} & Discrete & 1024 & 17 & 22.2 & 0.201 & 0.703 & \phantom{0}509.0 \\
        PVDM-AE \citep{yu2023video} & Continuous & 1152 & 16 & 19.1 & 0.333 & 0.563 & 1270.1\\
        OmniTokenizer-CV \citep{wang2024omnitokenizer} & Continuous & 1024 & 17 & 23.2 & 0.175 & 0.744 & \phantom{0}396.7 \\
        CosmosTokenizer-CV$^*$ \citep{agarwal2025cosmos} & Continuous & 1024 & 17 & 24.0 & 0.220 & 0.774 & \phantom{0}519.6 \\
        \bf{\sname} (Ours) & Continuous & 1280 & 128 & \bf{28.6} & \bf{0.066} & \bf{0.892} & \phantom{0}\bf{102.9} \\
        \bottomrule
    \end{tabular}\label{tab:main} 
\end{adjustbox}
\end{table*}

%% file: resource/generation_results_single_row.tex
\begin{figure*}[t]
\begin{minipage}{0.30\textwidth}
\centering\small
\captionof{table}{
\textbf{FVDs of video generation models} on the UCF-101 dataset (128-frame, 128$\times$128 resolution). $\downarrow$ indicates lower values are better.
}
\centering\small
\vspace{-0.10in}
\begin{adjustbox}{max width=\linewidth}
\begin{tabular}{lc}
\toprule
Method & FVD$\downarrow$ \\
\midrule
MoCoGAN \citep{tulyakov2018mocogan} & 3679.0 \\
+ StyleGAN2 \citep{karras2020analyzing} & 2311.3 \\
MoCoGAN-HD \citep{tian2021good} & 2606.5 \\
DIGAN \citep{yu2022generating} & 2293.7 \\
StyleGAN-V \citep{skorokhodov2022stylegan} & 1773.4 \\
PVDM-L \citep{yu2023video} & \phantom{0}505.0 \\
HVDM \citep{kim2024hybrid} & \phantom{0}549.7 \\
Latte-L/2 \citep{ma2024latte_rebuttal} & 1901.8 \\
\midrule
\sname-SiT-L/2 (Ours) & \phantom{0}\bf{369.3} \\
\bottomrule
\end{tabular}\label{tab:long-video-generation} 
\end{adjustbox}
\end{minipage}
\hfill
\begin{minipage}{0.35\textwidth}
\centering\small
\captionof{table}{
\textbf{Video generation efficiency.} We report time (s) and memory (GB) required for synthesizing a 128-frame video using a single NVIDIA 4090 24GB GPU.
We use the DDIM sampler \citep{song2020denoising} with 200 sampling steps for PVDM-L and HVDM and use the Euler-Maruyama sampler \citep{ma2024sit} with 250 sampling steps for our method.
}
\begin{adjustbox}{max width=\linewidth}
\begin{tabular}{lcc}
\toprule
Method  & Time (s) & Mem (GB) \\
\midrule
TATS \citep{ge2022long} &  180.7 & 9.8 \\
LARP \citep{wang2024larp} & 114.3 & \bf{3.1} \\
\midrule
PVDM-L \citep{yu2023video} & 116.9 & 4.0 \\
HVDM \citep{kim2024hybrid} & \phantom{0}52.1 & 3.9 \\
Latte-L/2 \citep{ma2024latte_rebuttal} & \phantom{0}21.4 & \textbf{3.1} \\
\midrule
\sname-SiT-L/2 (Ours) & \phantom{0}\phantom{0}\textbf{9.8} & 4.5 \\
\bottomrule
\end{tabular}\label{tab:long-video-generation-efficiency} 
\end{adjustbox}
\end{minipage}
\hfill
\begin{minipage}{0.32\textwidth}
\vspace{-0.05in}
\centering
    \includegraphics[width=0.95\linewidth]{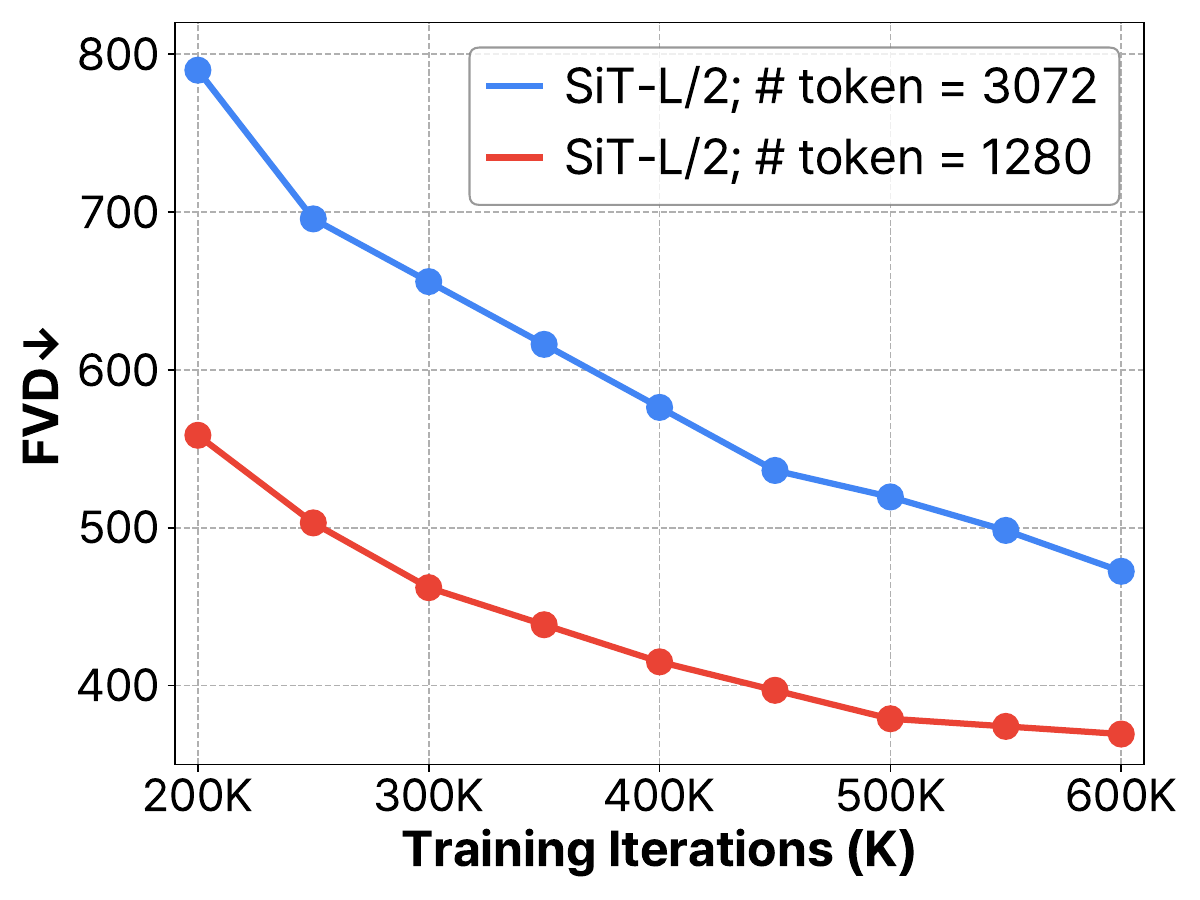}
    \vspace{-0.1in}
    \captionof{figure}{
    \textbf{Efficient video tokenization improves video generation.}
    We report FVDs of SiT-L/2 models trained upon CoordTok with token sizes of 1280 and 3072. $\downarrow$ indicates lower values are better.
    }
    \label{fig:generation_fvd_curve}
\end{minipage}
\end{figure*}

%% file: resource/analysis_scalability.tex
\begin{figure*}[tb]
\centering
\begin{subfigure}{.3\textwidth}
\centering
\includegraphics[width=0.9\textwidth]{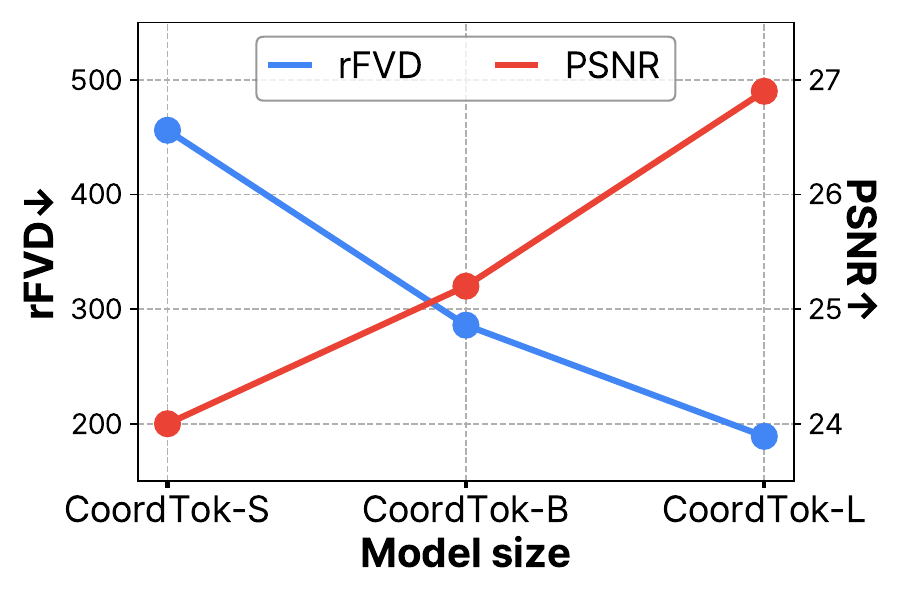}
\vspace{-0.05in}
\caption{Effect of Model size} 
\label{subfig:analysis-model-size}
\end{subfigure}
~~
\begin{subfigure}{.3\textwidth}
\centering
\includegraphics[width=0.9\textwidth]{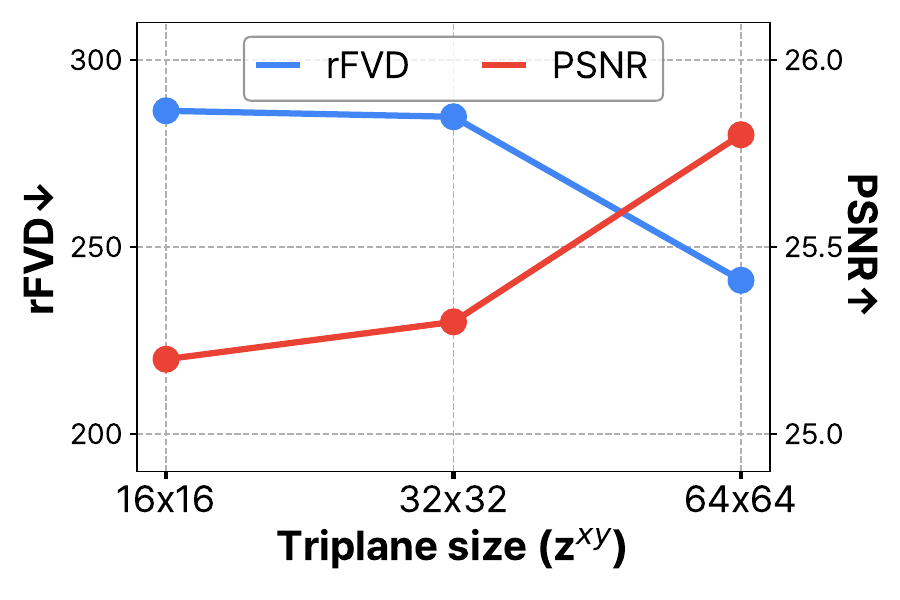}
\vspace{-0.05in}
\caption{Effect of Triplane size (spatial)} 
\label{subfig:analysis-spatial}
\end{subfigure}
~~
\begin{subfigure}{.3\textwidth}
\centering
\includegraphics[width=0.9\textwidth]{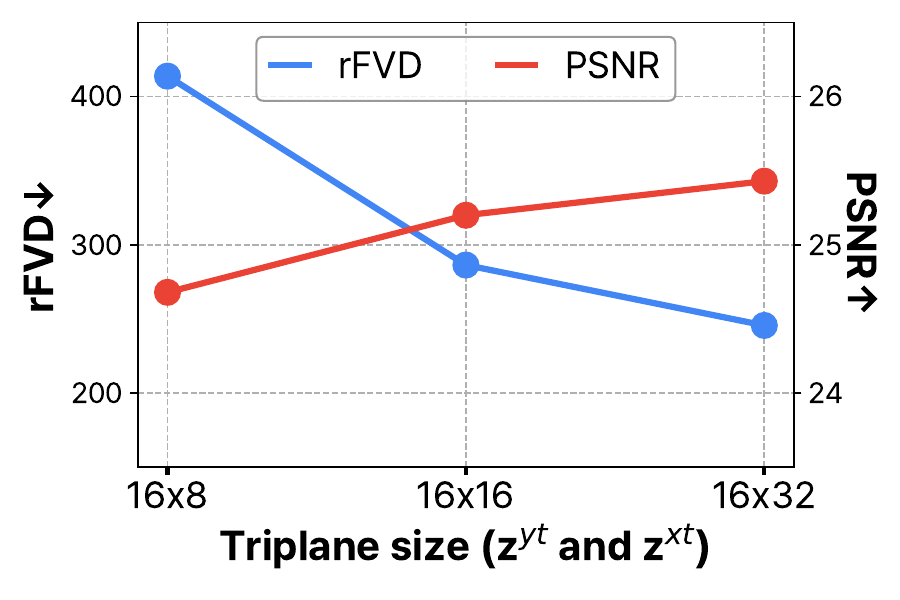}
\vspace{-0.05in}
\caption{Effect of Triplane size (temporal)}
\label{subfig:analysis-temporal}
\end{subfigure}
\vspace{-0.1in}
\caption{\textbf{Analysis on the effect of (a) model size, (b) spatial dimensions of triplane representations, and (c) temporal dimensions of triplane representations.}
For our main experiments, we use CoordTok-L with triplane representations of 16$\times$16 spatial dimensions and 32 temporal dimensions.
$\downarrow$ and $\uparrow$ denote whether lower or higher values are better, respectively.
}
\label{fig:analysis-scalability}
\end{figure*}

%% file: resource/analysis_recon_vs.tex
\begin{figure*}[tb]
\centering
\begin{subfigure}{.45\textwidth}
\centering
\includegraphics[width=0.8\textwidth]{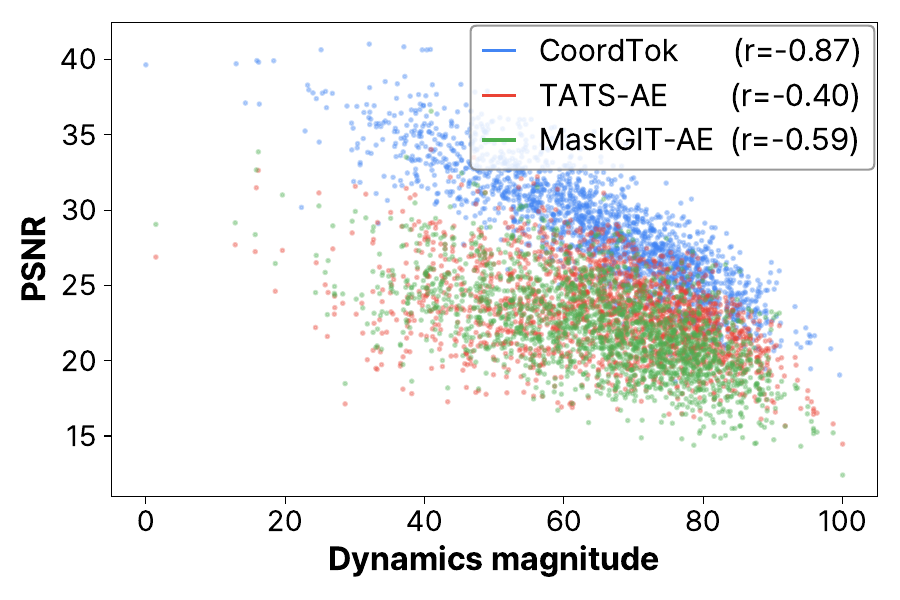}
\caption{{Effect of triplane representations.}} 
\label{subfig:recon_frame_distance}
\end{subfigure}
\begin{subfigure}{.45\textwidth}
\centering
\includegraphics[width=0.8\textwidth]{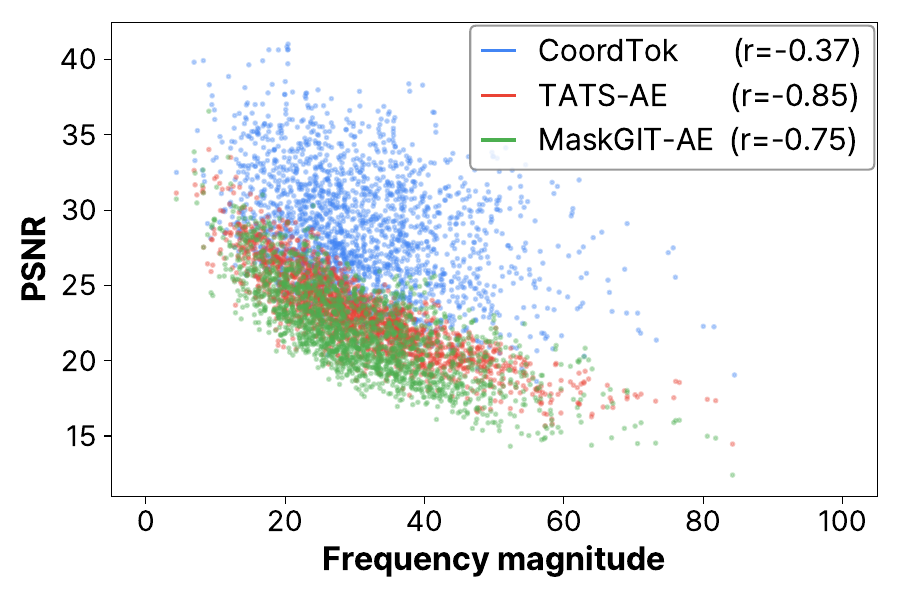}
\caption{{Effect of coordinate-based representations}}
\label{subfig:recon_frequency}
\end{subfigure}
\vspace{-0.07in}
\caption{
\textbf{Analysis on the effect of (a) triplane representations and (b) coordinate-based representations.}
(a) We measure the Pearson correlation $r$ between the reconstruction quality and a dynamics metric that measures how dynamic each video is. A video with a larger dynamics magnitude indicates a more dynamic video.
We find that the correlation is stronger for CoordTok compared to TATS-AE \citep{ge2022long} and MaskGIT-AE \citep{chang2022maskgit}, which encode videos into 3D latents.
We hypothesize this is because it is difficult to decompose dynamic videos into contents ($\rvz^{xy}$) and motions ($\rvz^{yt}$, $\rvz^{xt}$).
(b) We measure the Pearson correlation $r$ between the reconstruction quality and a frequency metric that measures the fineness of video details \citep{yan2024elastictok}. A video with a larger frequency magnitude indicates a finer-grained video.
In this case, we find that the correlation is weaker for CoordTok compared to other tokenizers. We hypothesize this is because CoordTok explicitly learns a mapping from each coordinate-based representation to pixels within the corresponding patch.
}
\label{fig:recon_analysis}
\end{figure*} 

%% file: resource/analysis_partial_recon.tex
\begin{table}[t]
\caption{\textbf{Effect of sampling}.
We report rFVD and the maximum batch size (Max BS) measured with a single NVIDIA 4090 24GB GPU, with different sampling schemes.
Random patch uses center coordinates of randomly selected patches for training, 
while Random frame uses all coordinates from a few randomly sampled frames for training.
Ratio (\%) indicates the proportion of sampled coordinates relative to all possible coordinates within a video.
$\downarrow$ indicates lower values are better.
}\label{tab:analysis-partial-recon} 
\centering\small
\vspace{-0.07in}
\begin{tabular}{lccc}
\toprule
Sampling & Ratio (\%) &  rFVD$\downarrow$ & Max BS \\
\midrule
Random frame & 3.125 & 479 & 13\\\midrule
Random patch & 1.563 & 401 & 21 \\
Random patch & 3.125 & 238 & 13\\
\bottomrule
\end{tabular}
\end{table}

%% file: resource/latent_visualization.tex
\begin{figure*}[t]
\centering
\includegraphics[width=0.9\linewidth]{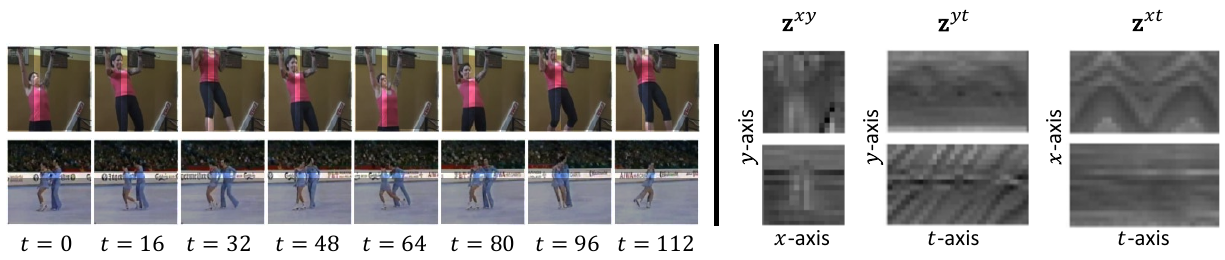}
\vspace{-0.1in}
\caption{
\textbf{Illustration of factorized triplane representations} $\rvz = [\rvz^{xy}, \rvz^{yt}, \rvz^{xt}]$ of CoordTok trained on the UCF-101 dataset \citep{soomro2012ucf101}.
We note that $\rvz^{xy}$ captures the global content in the video across time, \eg, layout and appearance of the scene or object, and $\rvz^{yt}$, $\rvz^{xt}$ capture the underlying motion in the video across two spatial axes.
}
\label{fig:latent-visualization}
\end{figure*}

%% file: sec/2_related_works.tex
\section{Related Work}
\paragraph{Video tokenization}
Many recent works have explored the idea of using video tokenizers to encode videos into low-dimensional latent tokens.
Initial attempts proposed to directly use image tokenizers for videos \citep{van2017neural,razavi2019generating,esser2021taming} via frame-wise compression.
However, this approach overlooks the temporal coherence of videos, resulting in inefficient compression.
Thus, recent works have proposed to train a tokenizer specialized for videos \citep{yan2021videogpt,ge2022long,hong2022cogvideo,blattmann2023align,yu2023magvit,yu2023language,yan2023temporally,yang2024cogvideox,videoworldsimulators2024,wang2024emu3,wang2024omnitokenizer}.
They typically extend image tokenizers by replacing spatial layers with spatiotemporal layers (\eg, 2D convolutional layers to 3D convolution layers).
More recent works have introduced efficient tokenization schemes with careful consideration of redundancy in video data.
For instance, several works proposed to encode videos into factorized triplane representations
\citep{yu2023video,kim2024hybrid,yu2024efficient}, and another line of works proposed an adaptive encoding scheme that utilizes the redundancy of videos for tokenization \citep{yan2024elastictok,wang2024larp}.
However, they still train the tokenizer through reconstruction of entire video frames, so training is only possible with short video clips split from the original long videos.
Our work introduces a video tokenizer that can directly handle much longer video clips by removing the need for a decoder to reconstruct entire video frames during training.
By capturing the global information present in long videos, we show that our tokenizer achieves more effective tokenization.

\paragraph{Latent video generation}

Instead of modeling distributions of complex and high-dimensional video pixels, most recent video generation models focus on learning the latent distribution induced by video tokenizers, as it can dramatically reduce memory and computation bottlenecks.
One approach involves training autoregressive models \citep{yan2021videogpt,kondratyuk2023videopoet,ge2022long} in a discrete token space \citep{van2017neural,razavi2019generating}.
Another line of research \citep{villegas2022phenaki,yan2023temporally, yoo2023towards} also considers discrete latent space but has trained masked generative transformer (MaskGiT; \citep{chang2022maskgit}) for generative modeling.
Finally, many recent works \citep{blattmann2023align,gupta2023photorealistic,yu2023video,kim2024hybrid,he2022lvdm,lu2023vdt,singer2022make,zhou2022magicvideo} have trained diffusion models \citep{sohl2015deep,ho2020denoising} in continuous latent space, inspired by the success of latent diffusion models in the image domain \citep{rombach2022high}.
Despite their efforts, the models are typically limited to processing only short video clips at a time (usually 16-frame clips), which makes it difficult for the model to generate longer videos.
In this paper, we significantly improve the limited contextual length of latent video generation models by introducing an efficient video tokenizer.

%% file: sec/5_conclusion.tex
\section{Conclusion}
In this paper, we have presented CoordTok, a scalable video tokenizer that learns a mapping from coordinate-based representations to the corresponding patches of input videos.
CoordTok is built upon our intuition that training a tokenizer directly on long videos would enable the tokenizer to leverage the temporal coherence of videos for efficient tokenization.
Our experiments show that CoordTok can encode long videos using far fewer number of tokens than existing baselines.
We also find that this efficient video tokenization enables memory-efficient training of video generation models that can generate long videos at once.
We hope that our work further facilitates future researches on designing scalable video tokenizers and efficient video generation models.

\paragraph{Limitations and future directions} One limitation of our work is that our tokenizer struggles more with dynamic videos than with static videos, as shown in \cref{fig:recon_analysis}.
We hypothesize this is due to the difficulty of learning to decompose dynamic videos into global content and motion.
One interesting future direction could involve introducing multiple content planes across the temporal dimension.
Moreover, future work may introduce an adaptive method for deciding the number of such content planes based on how dynamic each video is, similar to techniques in video codecs \citep{wiegand2003overview,sze2014high,mukherjee2015technical,han2021technical} or an adaptive encoding scheme designed for a recent video tokenizer \citep{yan2024elastictok}.
Lastly, we are excited about scaling up our tokenizer to longer videos from larger datasets and evaluating it on challenging downstream tasks such as long video understanding and generation.

%% file: sec/X_suppl.tex
\clearpage
\appendix
\setcounter{page}{1}
\maketitlesupplementary
\setcounter{section}{0}

\section{Implementation Details}
\label{appendix:implementation_details}
\subsection{Long video tokenization}
We train \sname via AdamW optimizer \citep{loshchilov2017decoupled} with a constant learning rate of $10^{-4}$, $(\beta_1, \beta_2)=(0.9, 0.999)$, and weight decay $0.001$. We use a batch size of 256, where each sample is a randomly sampled 128-frame video. \sname is trained in two stages: main training and fine-tuning. In the main training stage, we reconstruct $N=1024$ randomly sampled coordinates and update the model using $\ell_2$ loss. In the fine-tuning stage, we reconstruct 16 randomly sampled frames (i.e., $N=4096$ coordinates) and update the model using a combination of $\ell_2$ loss and LPIPS loss with equal weights.
To speed up training, we use mixed-precision (fp16). For the main experimental results, we train \sname for 1M iterations and fine-tune it for 50K iterations. For analysis and ablation studies, we train \sname for 200K iterations and fine-tune it for 10K iterations.

\paragraph{Architecture}
CoordTok consists of a \textit{transformer encoder} that extracts video features from raw videos, a \textit{cross-self encoder} that processes video features into triplane representations via cross-attention between learnable parameters and video features, and a \textit{transformer decoder} that learns a mapping from coordinate-based representations into corresponding patches. 
In what follows, we describe each component in detail.
\begin{itemize}
    \item \textbf{Transformer encoder} consists of a Conv3D patch embedding, learnable positional embedding, and transformer layers, where each transformer layer comprises self-attention and feed-forward layers.
    \item \textbf{Cross-self encoder} consists of plane-wise Conv2D patch embeddings, transformer layers, and plane-wise linear projectors, where each transformer layer comprises cross-attention, self-attention, and feed-forward layers.
    \item \textbf{Transformer decoder} consists of linear patch embedding, learnable positional embedding, transformer layers, and a linear projector, where each transformer layer comprises self-attention and feed-forward layers.
\end{itemize}

We provide the detailed architecture configurations for each model size in \Cref{tab:model-config}.
\input{resource/appendix_model_config}

\subsection{Long video generation}
We implement CoordTok-SiT-L/2 based on the original SiT implementation \citep{ma2024sit}.
The inputs of SiT-L/2 are the normalized triplane representation obtained by tokenizing video clips of length 128 with CoordTok. To normalize the triplane representation, we randomly sample 2048 video clips of length 128 and calculate the mean and standard deviation for each plane. We train SiT-L/2 via AdamW optimizer \citep{loshchilov2017decoupled} with a constant learning rate of $10^{-4}$, $(\beta_1, \beta_2)=(0.9, 0.999)$, and no weight decay. We use a batch size of 64. We train the model for 600K iterations and we update an EMA model with a momentum parameter 0.9999.

\paragraph{Architecture} We use the same structure as SiT, except that our patch embedding and final projection layers are implemented separately for each plane.
To train the unconditional video generation model, we assume the number of classes as 1, and we set the class dropout ratio to 0. We provide the detailed architecture configurations in \Cref{tab:architecture_sit}.
\input{resource/appendix_model_architecture_sit}

\paragraph{Sampling} For sampling, we use the Euler-Maruyama sampler with 250 sampling steps and a diffusion coefficient $w_t=\sigma_t$. We use the last step of the SDE sampler as 0.04.

\section{Evaluation Details}
\label{appendix:evaluation_details}
\subsection{Long video reconstruction}
For our \sname, we tokenize and reconstruct 128-frame videos all at once. Specifically, we encode the video into a triplane representation and then reconstruct the video by passing all patch coordinates through the transformer decoder at once. In contrast, the baselines can only handle videos of much shorter lengths (e.g., 16 frames for PVDM-AE \citep{yu2023video}). Therefore, to evaluate the reconstruction quality of 128-frame videos for the baselines, we split the videos into short clips and tokenize and reconstruct them. To be specific, we first split a 128-frame video into shorter clips suitable for each tokenizer. We then tokenize and reconstruct each short clip individually using the tokenizer. Finally, we concatenate all the reconstructed short clips to obtain the 128-frame video.

For evaluating the quality of reconstructed videos, we follow the setup of MAGVIT \citep{yu2023magvit}. We randomly sample 10000 video clips of length 128, and then measure the reconstruction quality using the metrics as follows:
\begin{itemize}
    \item \textbf{rFVD} \citep{unterthiner2019fvd} measures the feature distance between the distributions of real and reconstructed videos. It uses the I3D network \citep{carreira2017quo} to extract features, and it computes the distance based on the assumption that both feature distributions are multivariate Gaussian. Specifically, we compute the rFVD score on video clips of length 128.
    \item \textbf{PSNR} measures the similarity between pixel values of real and reconstructed images using the mean squared error. For videos, we compute the PSNR score for each frame and then average these frame-wise PSNR scores. 
    \item \textbf{LPIPS} \citep{zhang2018unreasonable} measures the perceptual similarity between real and reconstructed images by computing the feature distance using a pre-trained VGG network \citep{simonyan2014very}. It aggregates the distance of features extracted from various layers. For videos, we compute the LPIPS score for each frame and then average these frame-wise LPIPS scores.
    \item \textbf{SSIM} \citep{wang2004image} measures the structural similarity between real and reconstructed images by comparing luminance, contrast, and structural information. For videos, we compute the SSIM score for each frame and then average these frame-wise SSIM scores.
\end{itemize}

\subsection{Long video generation}
For sname-SiT-L/2, we generate the tokens corresponding to a 128-frame video all at once and then decode these tokens using \sname. In contrast, baselines iteratively generate 128-frame videos. For instance, PVDM and HVDM generate the next 16-frame video conditioned on the previously generated 16-frame video clip.

For evaluating the quality of generated videos, we strictly follow the setup of StyleGAN-V \citep{skorokhodov2022stylegan} that calculates the FVD scores \citep{unterthiner2019fvd} between the distribution of real and generated videos. To be specific, we use 2048 video clips of length 128 for each distribution, where the real videos are sampled from the dataset used to train generation models (\ie, the UCF-101 dataset \citep{soomro2012ucf101}).

\subsection{Analysis}

\begin{itemize}
    \item \textbf{Dynamics magnitude} To measure how dynamic each video is, we use the pixel value differences between consecutive frames. To be specific, we compute the dynamics magnitude for each pair of consecutive frames, calculate the mean of these values, and then take the logarithm. Here, dynamics magnitude of two frames $f^1$ and $f^2$ of resolution $H \times W$ can be defined as follows:
\begin{align}\label{eq:frame_distance}
d(f^1, f^2) ={\frac{1}{HW}} \sum_{h=1}^H \sum_{w=1}^W d_{2} (f^1_{hw}, f^2_{hw}),
\end{align}
where $f^i_{hw}$ denotes the RGB values at coordinates $(h, w)$ of frame $f^i$ and $d_2$ denotes $\ell_2$-distance of RGB pixel values. In \Cref{subfig:recon_frame_distance}, we standardize the video dynamics score into a range of 0 to 100.
    \item \textbf{Frequency magnitude} To measure the frequency magnitude, we use the metric proposed in \citet{yan2024elastictok} that utilizes a Sobel edge detection filter. To be specific, to get the frequency magnitude, we apply both horizontal and vertical Sobel filters to each frame to compute the gradient magnitude at each pixel. We then calculate the average of these magnitudes across all pixels.
\end{itemize}

\section{Baselines}
\label{appendix:baselines}

\subsection{Long video reconstruction}
We describe the main idea of baseline methods that we used for the evaluation. We also provide the shape of tokens of baselines in \Cref{tab:baseline_token_shape}.

\begin{itemize}
    \item \textbf{MaskGiT-AE} \citep{chang2022maskgit} uses 2D VQ-GAN \citep{esser2021taming} that encodes an image into a 2D discrete tokens.
    \item \textbf{TATS-AE} \citep{ge2022long} introduces 3D-VQGAN that compresses a 16-frame video clip both temporally and spatially into 3D discrete tokens.
    \item \textbf{MAGVIT-AE-L} \citep{yu2023magvit} also introduces 3D-VQGAN but improves architecture design (\eg, uses deeper 3D discriminator rather than two shallow discriminators for 2D and 3D separately, uses group normalization \citep{wu2018group} and Swish activation \citep{ramachandran2017searching}) and scales up the model size.
    \item \textbf{PVDM-AE} \citep{yu2023video} encodes a 16-frame video clip into factorized triplane representations.
    \item \textbf{LARP} \citep{wang2024larp} encodes videos into 1D arrays by utilizing a next-token prediction model as a prior model.
    \item \textbf{OmniTokenizer-DV} \citep{wang2024omnitokenizer} introduces image-video joint VQGAN that compresses a 17-frame video clip into 3D discrete tokens with more advanced architecture design (\eg, uses both 2D and 3D patch embedding layers to support both image and video tokenization, uses transformer backbone with causal attention layers).
    \item \textbf{OmniTokenizer-CV} \citep{wang2024omnitokenizer} uses the same architecture design as OmniTokenizer-DV, but replaces the VQ loss with KL loss so that it compresses a 17-frame video clip into 3D continuous latent vectors.
\end{itemize}
\input{resource/appendix_baseline_token_shape}

\subsection{Long video generation}
We describe the main idea of baseline methods that we used for the evaluation.
\begin{itemize}
    \item \textbf{MoCoGAN} \citep{tulyakov2018mocogan} proposes a video generative adversarial network (GAN; \citep{goodfellow2014generative}) that has a separate content generator and an autoregressive motion generator for generating videos.
    \item \textbf{MoCoGAN-HD} \citep{tian2021good} also proposes a video GAN with motion-content decomposition but uses a strong pre-trained image generator (StyleGAN2 \citep{karras2020analyzing}) for a high-resolution image synthesis.
    \item \textbf{DIGAN} \citep{yu2022generating} interprets videos as implicit neural representation (INR; \citep{sitzmann2020implicit}) and trains GANs to generate such INR parameters.
    \item \textbf{StyleGAN-V} \citep{skorokhodov2022stylegan} also introduces an INR-based video GAN with a computation-efficient discriminator.
    \item \textbf{PVDM-L} \citep{yu2023video} proposes a latent video diffusion model that generates videos in a projected triplane latent space.
    \item \textbf{HVDM} \citep{kim2024hybrid} proposes a latent video diffusion model that generates videos with 2D triplane and 3D wavelet representation.
    \item \textbf{Latte-L/2} \citep{ma2024latte_rebuttal} proposes a latent video diffusion transformer that generates video by processing latent vectors with alternating spatial and temporal attention layers.
\end{itemize}

\section{Additional Analysis}\label{appendix:additional_analysis}
\paragraph{Computational costs}\label{appendix:computational_costs}
We provide the GPU memory usage during training in \Cref{subfig:long-scalability}, and FLOPs during training in \Cref{fig:appendix:flops}. We find that our decoder design allows the efficient long video tokenization in terms of both GPU memory and FLOPs.

\input{resource/appendix_flops}

\paragraph{Analysis on the number of tokens}
We provide the reconstruction quality of CoordTok with 1280 and 3072 tokens in \Cref{tab:appendix_recon_n_token}. Although there is no significant difference in the reconstruction quality between CoordTok with token sizes of 1280 and 3072, training SiT-L/2 with the 1280 tokens results in substantially better generation quality (see \Cref{sec:long_video_generation}).
\input{resource/appendix_recon_numtokens}

\paragraph{Analysis on the effect of LPIPS fine-tuning}
In \Cref{tab:appendix_effect_2step}, we investigate the effect of the additional fine-tuning phase, where we train \sname with both $\ell_2$ loss and LPIPS loss \citep{zhang2018unreasonable} for 50K iterations after training \sname with $\ell_2$ loss for 1M iterations. We find that fine-tuning phase improves the perceptual quality (\ie, rFVD score: 188.3 $\rightarrow$ 102.9, and LPIPS score: 0.141 $\rightarrow$ 0.066), but degrades the pixel-level reconstruction quality (\ie, PSNR: 30.3 $\rightarrow$ 28.6, and SSIM: 0.905 $\rightarrow$ 0.892).

\input{resource/appendix_effect_2step}

\section{Additional Quantitative Results}
\label{appendix:additional_quantitative_results}

\paragraph{16-frame reconstruction quality}
To further evaluate the quality of reconstructed videos from tokenizers, we report the rFVD score on video clips of length 16 for the \sname and other tokenizers with varying number of token sizes in \Cref{fig:appendix:fvd16}. For evaluation, we use 10000 video clips of length 128, which are also used to measure the rFVD score on 128-frame videos. We split each 128-frame video into 16 non-overlapping sub-clips, and then compute the rFVD score on totally 80000 video clips of length 16.

\input{resource/appendix_fvd16}

\section{Additional Qualitative Results}
\label{appendix:additional_qualitative_results}
In \Cref{fig:appendix_qual_recon}, we provide additional video reconstruction results from \sname. In addition, in \Cref{fig:appendix_qual_gen,fig:appendix_qual_gen2}, we provide unconditional video generation results from \sname-SiT-L/2.

\input{resource/appendix_qualitative_reconstruction}
\input{resource/appendix_qualitative_generation}

%% file: resource/appendix_model_config.tex
\begin{table}[ht]
\caption{Model configurations of \sname for each model size.}
\label{tab:model-config} 
\centering\small
\vspace{-0.05in}
\begin{adjustbox}{max width=\linewidth}
\begin{tabular}{llccc}
\toprule
Model size & Module & \#layers & Hidden dim. & \#heads \\
\midrule
\multirow{3}{*}{Large} & Transformer Encoder & 8 & 1024 & 16\\
 & Cross-self Encoder & 24 & 1024 & 16\\
 & Transformer Decoder & 24 & 1024 & 16 \\
\midrule
\multirow{3}{*}{Base} & Transformer Encoder & 8 & 768 & 12\\
 & Cross-self Encoder & 12 & 768 & 12\\
 & Transformer Decoder &12 & 768 & 12 \\
\midrule
\multirow{3}{*}{Small} & Transformer Encoder & 8 & 512 & 8\\
 & Cross-self Encoder & 8 & 512 & 8\\
 & Transformer Decoder & 8 & 512 & 8 \\
\bottomrule
\end{tabular}
\end{adjustbox}
\end{table}

%% file: resource/appendix_model_architecture_sit.tex
\begin{table}[h]
\caption{Model configurations of CoordTok-SiT-L/2.}
\label{tab:architecture_sit} 
\centering\small
\vspace{-0.05in}
\begin{adjustbox}{max width=\linewidth}
\begin{tabular}{lcc}
\toprule
& SiT-L/2, \#token = 1280 & SiT-L/2, \#token = 3072 \\
\midrule
Input dim. ($\rvz^{xy}$) & 16$\times$16$\times$8 & 32$\times$32$\times$8\\
Input dim. ($\rvz^{yt}$) & 16$\times$32$\times$8 & 32$\times$32$\times$8\\
Input dim. ($\rvz^{xt}$) & 16$\times$32$\times$8 & 32$\times$32$\times$8\\
\midrule
\# layers & 24 & 24\\
Hidden dim. & 1024 & 1024\\
\# heads & 16 & 16\\
\bottomrule
\end{tabular}
\end{adjustbox}
\end{table}

%% file: resource/appendix_baseline_token_shape.tex
\begin{table}[ht]
\caption{Token shapes of video tokenization baselines}
\label{tab:baseline_token_shape} 
\centering\small
\vspace{-0.05in}
\begin{adjustbox}{max width=\linewidth}
\begin{tabular}{lcc}
\toprule
Method & Input shape & Token shape \\
\midrule
MaskGiT-AE \citep{chang2022maskgit} & 128$\times$128$\times$3 & 8$\times$8 \\
\midrule
TATS-AE \citep{ge2022long} & 16$\times$128$\times$128$\times$3 & 4$\times$16$\times$16\\
MAGVIT-AE-L \citep{yu2023magvit} & 16$\times$128$\times$128$\times$3 & 4$\times$16$\times$16\\
PVDM-AE \citep{yu2023video} & 16$\times$128$\times$128$\times$3 & (16$\times$16) $\times$ 3 \\
LARP \citep{wang2024larp} & 16$\times$128$\times$128$\times$3 & 1024 \\
OmniTokenizer \citep{wang2024omnitokenizer} & (1+16)$\times$128$\times$128$\times$3 & (1+4)$\times$16$\times$16 \\
\bottomrule
\end{tabular}
\end{adjustbox}
\end{table}

%% file: resource/appendix_flops.tex
\begin{figure}[h]
\vspace{-0.01in}
\centering\small
    \includegraphics[width=0.9\linewidth]{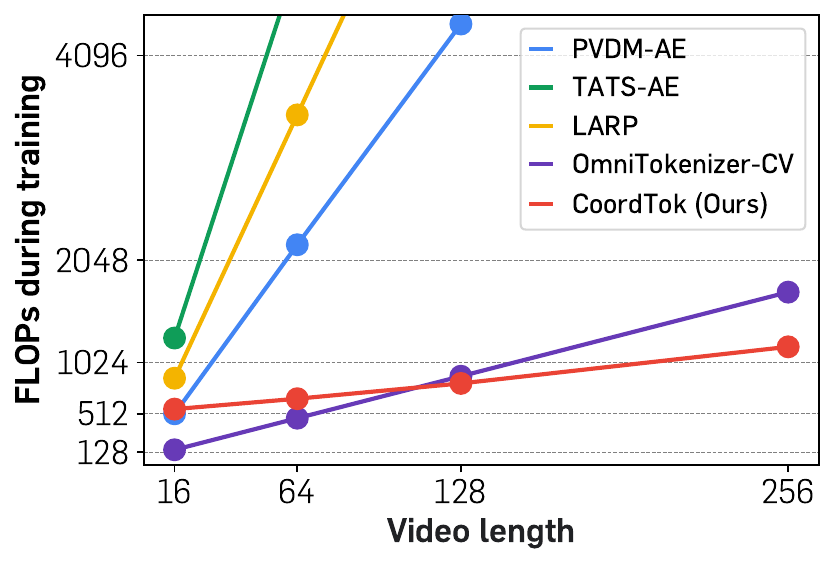}
    \vspace{-0.15in}
    \caption{FLOPs when training video tokenizers on 128$\times$128 resolution videos with varying lengths.
    }
    \label{fig:appendix:flops}
\end{figure}

%% file: resource/appendix_recon_numtokens.tex
\begin{table}[h]
\caption{Reconstruction quality of \sname with varying number of token sizes, evaluated on 128-frame videos. $\downarrow$ and $\uparrow$ denotes whether lower or higher values are better, respectively.}
\label{tab:appendix_recon_n_token} 
\centering\small
\vspace{-0.05in}
\begin{adjustbox}{max width=\linewidth}
\begin{tabular}{lcccc}
\toprule
\#tokens & rFVD$\downarrow$ & PSNR$\uparrow$ & LPIPS$\downarrow$ & SSIM$\uparrow$ \\
\midrule
1280 & 102.9 & 28.6 & 0.066 & 0.892 \\
3072 & \bf{100.5} & \bf{28.7} & \bf{0.065} & \bf{0.894} \\
\bottomrule
\end{tabular}
\end{adjustbox}
\end{table}

%% file: resource/appendix_effect_2step.tex
\begin{table}[h]
\caption{Effect of LPIPS fine-tuning phase for \sname. $\downarrow$ and $\uparrow$ denotes whether lower or higher values are better, respectively.}
\label{tab:appendix_effect_2step} 
\centering\small
\vspace{-0.05in}
\begin{adjustbox}{max width=\linewidth}
\begin{tabular}{llccccc}
\toprule
Phase & Iters &loss & rFVD$\downarrow$ & PSNR$\uparrow$ & LPIPS$\downarrow$ & SSIM$\uparrow$ \\
\midrule
1 &1M     & $\ell_2$ & 186.3 & \textbf{30.3} & 0.141 & \textbf{0.905} \\
2 & +50K  & $\ell_2+$LPIPS& \textbf{102.9} & 28.6 & \textbf{0.066} & 0.892 \\
\bottomrule
\end{tabular}
\end{adjustbox}
\end{table}

%% file: resource/appendix_fvd16.tex
\begin{figure}[h]
\vspace{-0.01in}
\centering\small
    \includegraphics[width=0.9\linewidth]{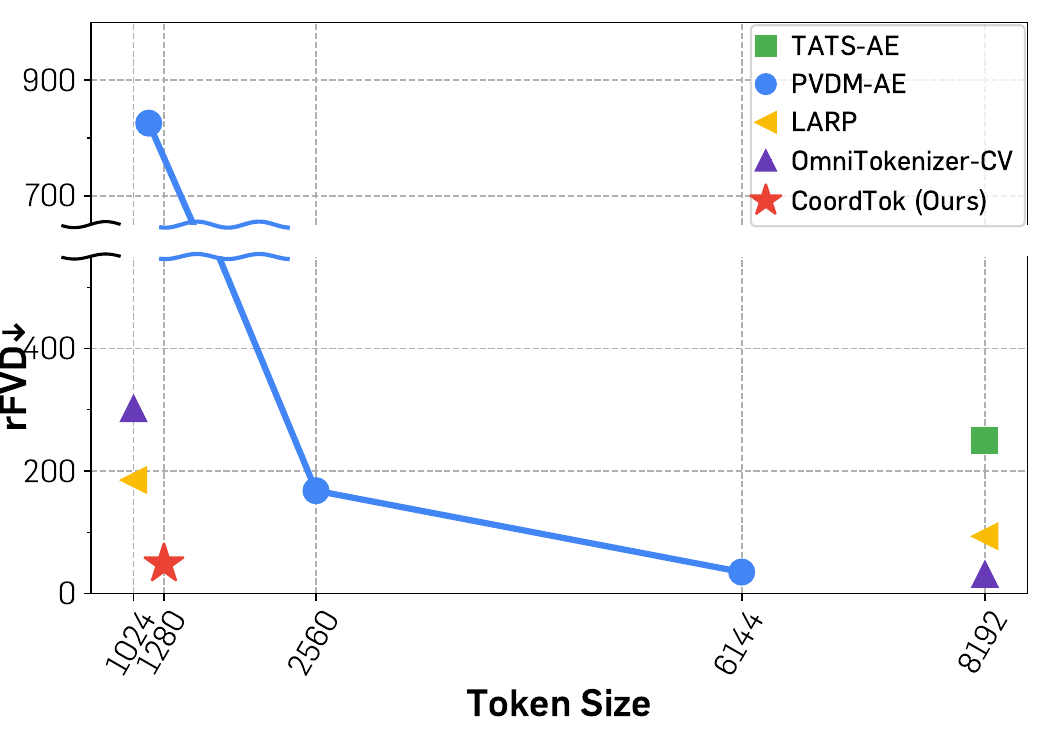}
    \vspace{-0.15in}
    \caption{rFVD scores of video tokenizers, evaluated on 16-frame videos, with respect to the token size used for encoding 128-frame videos. $\downarrow$ indicates lower values are better.
    }
    \label{fig:appendix:fvd16}
\end{figure}

%% file: resource/appendix_qualitative_reconstruction.tex
\begin{figure*}[ht]
    \includegraphics[width=\textwidth]{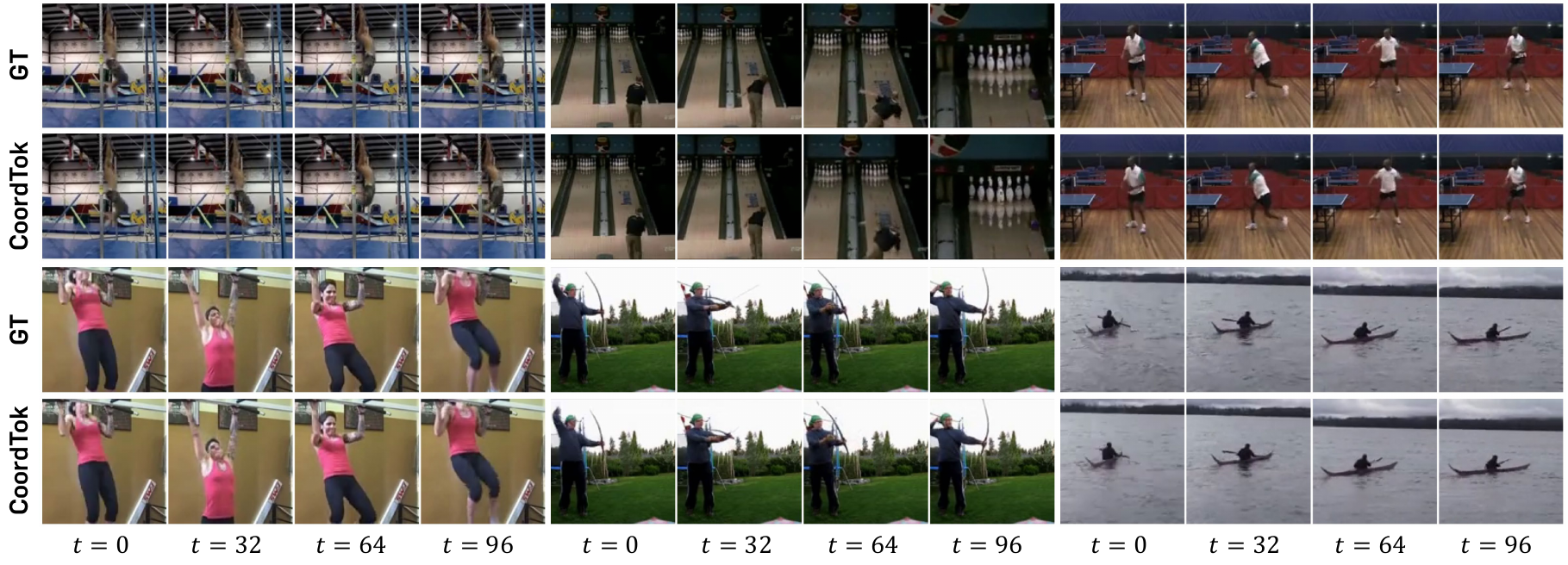}
    \caption{Additional 128-frame, 128$\times$128 resolution video reconstruction results from \sname (Ours) trained on the UCF-101 dataset \citep{soomro2012ucf101}. For each frame, we visualize the ground-truth (GT) and reconstructed pixels from \sname.}
    \label{fig:appendix_qual_recon}
\end{figure*}

%% file: resource/appendix_qualitative_generation.tex
\begin{figure*}[ht]
\centering
\includegraphics[width=0.95\linewidth]{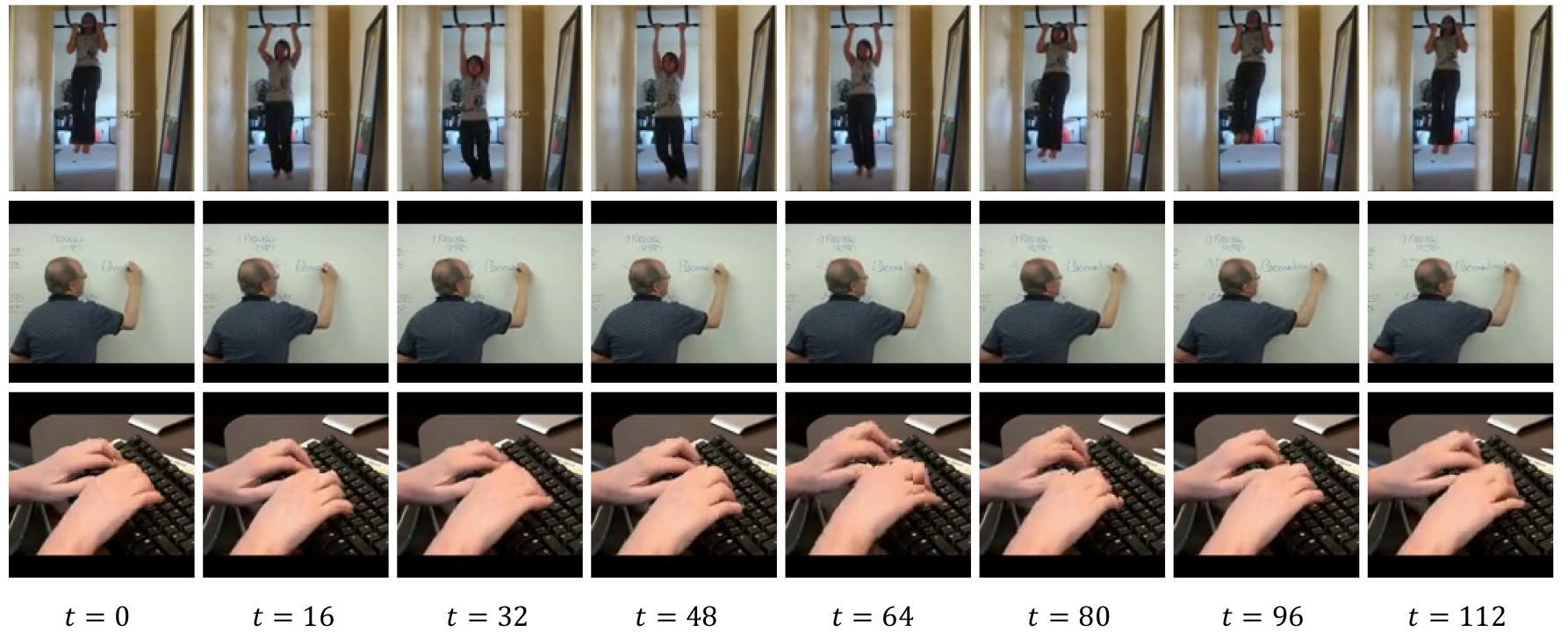}
\vspace{-0.1in}
\caption{Unconditional 128-frame, 128$\times$128 resolution video generation results from CoordTok-SiT-L/2 trained on 128-frame videos from the UCF-101 dataset \citep{soomro2012ucf101}.}
\label{fig:appendix_qual_gen}
\vspace{-0.1in}
\end{figure*}
\begin{figure*}[ht]
    \includegraphics[width=\textwidth]{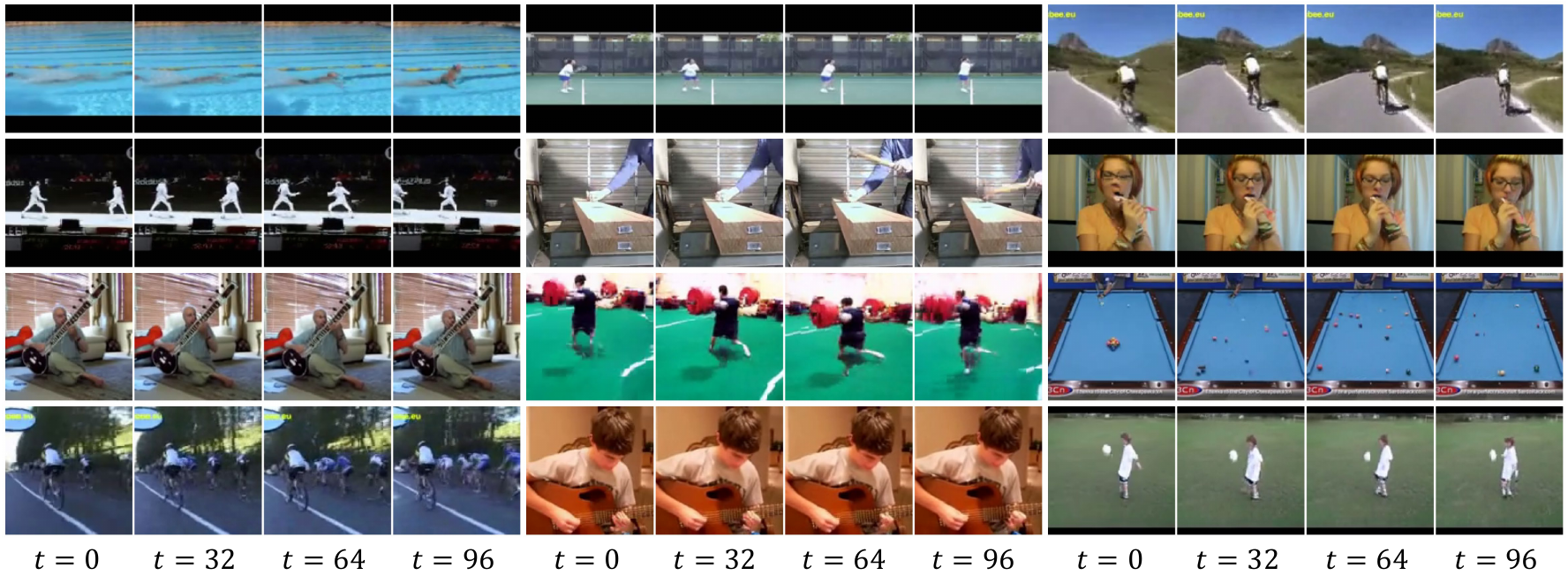}
    \caption{Unconditional 128-frame, 128$\times$128 resolution video generation results from \sname-SiT-L/2 trained on 128-frame videos from the UCF-101 dataset \citep{soomro2012ucf101}.}
    \label{fig:appendix_qual_gen2}
\end{figure*}